# Adversarially Robust Detection of Harmful Online Content: A Computational Design Science Approach

Yidong Chai

City University of Hong Kong

Email: yidong.chai@cityu.edu.hk

Yi Liu

Hefei University of Technology

Email: yiliu@mail.hfut.edu.cn

Mohammadreza Ebrahimi

University of South Florida

Email: ebrahimim@usf.edu

Weifeng Li

University of Georgia

Email: weifeng.li@uga.edu

Balaji Padmanabhan

University of Maryland

Email: bpadmana@umd.edu

# Adversarially Robust Detection of Harmful Online Content: A Computational Design Science Approach

## Abstract

Social media platforms are plagued by harmful content such as hate speech, misinformation, and extremist rhetoric. Machine learning (ML) models are widely adopted to detect such content; however, they remain highly vulnerable to adversarial attacks, wherein malicious users subtly modify text to evade detection. Enhancing adversarial robustness is therefore essential, requiring detectors that can defend against diverse attacks (generalizability) while maintaining high overall accuracy. However, simultaneously achieving both optimal generalizability and accuracy is challenging. Following the computational design science paradigm, this study takes a sequential approach that first proposes a novel framework (Large Language Model-based Sample Generation and Aggregation, LLM-SGA) by identifying the key invariances of textual adversarial attacks and leveraging them to ensure that a detector instantiated within the framework has strong generalizability. Second, we instantiate our detector (Adversarially Robust Harmful Online Content Detector, ARHOCD) with three novel design components to improve detection accuracy: (1) an ensemble of multiple base detectors that exploits their complementary strengths; (2) a novel weight assignment method that dynamically adjusts weights based on each sample's predictability and each base detector's capability, with weights initialized using domain knowledge and updated via Bayesian inference; and (3) a novel adversarial training strategy that iteratively optimizes both the base detectors and the weight assignor. We addressed several limitations of existing adversarial robustness enhancement research and empirically evaluated ARHOCD across three datasets spanning hate speech, rumor, and extremist content. Results show that ARHOCD offers strong generalizability and improves detection accuracy under adversarial conditions. Our study contributes to IS research at the intersection of social media management, AI security, and computational design science, and offers practical implications for users, platforms, and regulators.

## 1. Introduction

Social media platforms enable people to share experiences and ideas, playing a key role in information dissemination and public discourse. However, they are also inundated with harmful content, including hate speech, misinformation, and extremist rhetoric. Such content disrupts users' lives, fuels social polarization, and threatens societal stability. Reports note that "social media allows extremist ideologies, conspiracies, dis- and misinformation to be shared at an unprecedented scale and speed", becoming "the world's most potent incubator of extremism"[1]. Detecting such harmful content is therefore essential. Given the massive volume of online content and the strengths of Machine Learning (ML) models in text analysis, ML models have been widely adopted for automated harmful content detection (Etudo and Yoon 2024, Wei et al. 2022).

Harmful content is often crafted by users with goals such as spreading misinformation, promoting extremism, or inciting hatred. To achieve their goals, these users are motivated to evade detection. This challenge is exacerbated by the vulnerabilities of ML models. Although ML models can achieve high accuracy, they are vulnerable to adversarial manipulation. In harmful content detection, once a harmful text is flagged, a user can slightly modify its wording without altering its meaning, to create a new version that deceives the detector into misclassifying it as non-harmful. The modified text is an *adversarial sample*, and the initial text is a *clean sample*. The process where a detector initially detects harmful content but fails after subtle wording modifications is known as an *adversarial attack*. Adversarial attacks have been shown to fool various detectors. For instance, Oak and Berkeley (2019) showed that the statement "go back to where you came from these fu**ing immigrants are destroying America" is correctly detected as hate, but when "fu**ing immigrants" is changed to "fuking immgrants," the classifier misclassifies it as non-hate, even though the edits are superficial and the meaning remains unchanged. Hence, there is a pressing need to enhance detectors' ability to withstand such attacks. Such ability is *adversarial robustness*. Enhancing adversarial robustness involves two key requirements. First, since many attack methods already exist and new ones continue to emerge, a detector must handle a broad range of attacks, including new ones, to ensure strong worst-case performance, which we refer to as *high generalizability*. Second, because real-world attacks can vary widely, detectors must also maintain strong average performance, which is *high accuracy*. Robust detectors must therefore perform well in both worst-case scenarios and typical real-world settings.

Adversarial attacks affect not only text-based ML models but also models in other fields, such as computer vision. Many methods have been proposed to enhance adversarial robustness across these fields. However, these enhancement methods still have major drawbacks that limit both the generalizability and accuracy of the detectors they enhance. A key challenge arises from the multi-tiered hierarchy of text, where characters form words and words form sentences. This hierarchical structure leads to diverse textual adversarial attacks, occurring at the character, word, or sentence level. However, existing methods typically focus on defending against attacks at only a specific level, failing to account for attacks at other levels. Consequently, detectors enhanced by these methods may perform well against certain attacks (e.g., achieving 90% accuracy) but poorly against others (e.g., only 10% accuracy). Moreover, adversarial attacks evolve continuously, creating an ongoing arms race between attackers and defenders. The hierarchical nature of text compounds this problem: not only can attacks take various known forms, but new ones can also emerge at any level. As a result, detectors robust to known attacks may still be misled by new ones.

This means that detectors robustified by existing methods exhibit *limited generalizability* across both known and new attacks. In addition, the existing robustness enhancement methods also struggle to deliver high accuracy for the detectors they enhance (i.e., *limited accuracy*, will be discussed in later content).

Achieving both optimal generalizability and accuracy is challenging because (1) each detector is a discrete ML model, leading to a discrete search space that is difficult to optimize, (2) potential attacks are vast and potentially unbounded, particularly due to the hierarchical nature of text, and (3) the two objectives may conflict (e.g., overfitting to common attacks may improve accuracy but reduce generalizability). To address the first challenge, we adopt the computational design science paradigm. Rather than directly optimizing the discrete search space, we focus on designing a novel detector that overcomes the limitations of existing methods. Since the ideal detector would achieve optimal performance, the observed improvements indicate that our design moves closer to the ideal detector and better addresses the underlying optimization problem. To tackle the second and third challenges, we employ a sequential approach where we first propose a novel framework that ensures high generalizability for detectors instantiated within it, and then instantiate a detector with multiple novel design components to further improve overall accuracy.

Specifically, we first propose a <u>L</u>arge <u>L</u>anguage <u>M</u>odel-based <u>S</u>ample <u>G</u>eneration and <u>A</u>ggregation (LLM-SGA) framework. The core idea is that, while textual adversarial attacks vary in form and occur at multiple levels, they share two key invariances: an adversarial sample and its initial clean sample convey the same meaning (*Invariance 1*), and the detector under attack has correctly detected the initial clean sample and generally performs well on diverse samples (*Invariance 2*). These invariances arise not only from the definition of adversarial attacks but also from the attacker's practical constraints. For Invariance 1, altering the meaning renders the attack ineffective, as the functional intent of the attack will be lost. For example, modifying "go back to where you came from these fu\*\*ing immigrants are destroying America" to "It's okay" would evade detection, but it no longer conveys the initial intent, so attackers would avoid such modifications. For Invariance 2, ML-based detectors typically perform quite well across diverse inputs. Moreover, if a clean sample is already misclassified as non-harmful, attackers have no incentive to modify it further, since the clean sample already achieves their goal. Based on the two invariances, our LLM-SGA proceeds in three steps. In Step 1, given an input sample, we prompt an LLM to generate multiple samples that preserve its meaning. In Step 2, the detector predicts both the input and the generated samples, yielding a predictive probability of harmfulness for each. In Step 3, these probabilities are aggregated, specifically averaged, to determine whether the input is harmful. Our LLM-SGA is generalizable for the following reason. If the input is adversarial, the generated samples will preserve its meaning, and thus also that of the clean sample (Invariance 1). Since the detector has correctly classified the clean sample and performs well on diverse samples (Invariance 2), it is expected to correctly classify the generated samples. While a detector can be misled by a single sample (e.g., adversarial one), incorporating multiple samples dilutes the effect of any adversarial one, thereby increasing the probability of correct detection. We also theoretically prove that this probability is lower-bounded, and the bound rises with the number of generated samples.

LLM-SGA is a generalizable framework that allows any instantiated detector to defend against various known and newly emerging adversarial attacks. However, the detection accuracy still depends on the specific detector instantiated within the framework. Designing such an accurate detector presents three key challenges, which motivate the three novel design components we introduce next to address them.

First, prior studies show that relying solely on one detector leaves detection systems vulnerable, as its weaknesses can be easily exploited to craft targeted attacks. To address this *single-detector challenge*, we adopt an ensemble approach, combining multiple detectors, each serving as a *base detector*, to leverage their complementary strengths. For each sample (initial input or generated ones), we obtain the predictive probability from each base detector. These probabilities form a two-dimensional matrix, which is then aggregated to produce the final result. We refer to this design component as the *two-dimensional ensemble* (Design Component 1). The novelty is that, unlike prior studies that aggregate along only one dimension, either samples or base models, we aggregate both, creating a novel two-dimensional ensemble structure.

Second, even among the samples that convey the same meaning, a detector's predictive probabilities may vary, making the probability of each sample being correctly detected differ. A prediction more likely to be correct is viewed as a *more accurate prediction*, and should receive a higher weight during aggregation. Since our first design component introduces multiple base detectors, the resulting two-dimensional probability matrix introduces additional complexity. Predictions vary not only across samples but also across base detectors. Assigning appropriate weights to each prediction, which we refer to as the *weighting challenge*, is therefore essential. To address this challenge, we propose a *Bayesian two-dimensional weight assignment method* (Design Component 2) to determine the weight of each prediction in the matrix. Since a prediction's accuracy depends on both samples and models (i.e., base detectors), we decompose it into a *sample's predictability* and a *model's capability*. We theoretically show that a sample's predictability is negatively correlated with the variance of its predictions across models, while a model's capability is negatively correlated with the variance of its predictions across samples. Based on these insights (as domain knowledge), we design novel variance-based priors for the weights to be assigned to each prediction, and then propose an attention-based neural network to infer the posterior weights. Since the posterior determines the final weights, we refer to this neural network as the *weight assignor*. The novelty lies in our Bayesian two-dimensional weighting mechanism, which integrates theoretically grounded variance-based priors with an attention-based network to derive posterior weights for each element in the two-dimensional matrix.

Third, unlike a single model whose parameters can be directly optimized by adversarial training (AT), our setting involves base detectors and a weight assignor that jointly determine detection outcomes, making it necessary to learn robust parameters for both. The data used to train the weight assignor should accurately capture the robustness of the base detectors. Using the same dataset for both base detectors and the weight assignor risks overfitting, causing the weight assignor to misallocate weights. Thus, the base detectors and the weight assignor need to be adversarially trained jointly but on separate datasets, known as the *learning challenge*. To address this, we propose an *iterative adversarial training strategy* (Design Component 3). We emulate attackers to generate adversarial samples, and then alternately train the base detectors with the weight assignor fixed, and train the weight assignor with the base detectors fixed. This alternating procedure enables the base detectors and the weight assignor to adapt to adversarial conditions while maintaining the required independence between their training data. The key novelty is an AT tailored for ensemble learning: unlike prior work that uses AT only to enhance base models, our strategy captures the interaction between the base models and the weight assignor, adversarially training both to enhance the ensemble's robustness.

We instantiate the three design components in the LLM-SGA framework, yielding our instantiated Adversarially Robust Harmful Online Content Detector (ARHOCD). We evaluated ARHOCD on three datasets covering hate speech, rumor, and extremist content. Two sets of experiments were conducted: the first evaluated generalizability, and the second evaluated accuracy against both known and new attacks. We further validated ARHOCD's advantages in broader settings, including cases where the base detectors are fixed and in an image classification task. We also conducted a case study to examine its advantages.

Our study contributes to interdisciplinary research on social media management, AI security, and computational design science. Detecting harmful content is crucial for a healthy social media environment and is a key topic in the IS field. This study highlights the security threats posed by adversarial attacks and focuses on strengthening the robustness of harmful content detection systems. To this end, we develop a general framework (LLM-SGA), a novel IT artifact that defends against diverse textual adversarial attacks by leveraging their underlying invariances. We instantiate this framework through three novel design components to develop ARHOCD, a situated implementation of an adversarially robust detector for harmful content detection. Beyond advancing harmful content detection, our study also informs broader AI security challenges, such as privacy protection, and provides a principled approach for enhancing adversarial robustness in other IS applications, including detecting hacker assets and identifying phishing attacks.

## 2. Research Background

### 2.1 Harmful Online Content Detection and Adversarial Attacks

The growing threat of harmful online content makes its detection imperative. As ML can learn the patterns pertaining to harmful content, it is commonly used for detection, with major studies shown in Appendix A. A major security threat to these detectors is adversarial attacks. According to Kurakin et al. (2017), an adversarial attack is defined as follows: given a clean sample $x^C$ with ground truth $y^C$ and a model $F$, the goal is to craft an adversarial sample $x^A$ that is perceptually indistinguishable from $x^C$ but causes $F$ to mispredict, despite $F$ correctly predicts the initial input $x^C$. Formally, $F(x^C) = y^C$, $F(x^A) \neq y^C$, and $x^C \sim x^A$, where we use $\sim$ to denote perceptual indistinguishability. For example, in computer vision where adversarial attacks were first observed, slight pixel perturbations to a panda image keep it perceptually unchanged, yet GoogLeNet shifts from correctly classifying it as a panda (57.7% predictive probability) to incorrectly classifying it as a gibbon (99.3% predictive probability) (Goodfellow et al. 2015).

Later studies showed that adversarial attacks also threaten models handling text, known as *textual adversarial attacks* (Papernot et al. 2016, Qiu et al. 2022). Text's multi-tiered hierarchy, where characters form words and words form sentences, complicates adversarial attacks and distinguishes them by enabling attacks at the character, word, or sentence level. Character-level attacks perturb text by inserting, flipping, swapping, or removing characters. For instance, in Table 1, attackers replace the characters s, a, and h in the word "trash" with ѕ (Cyrillic letter Dze), ɑ (Latin letter Alpha), and ʜ (Cherokee letter Ni). While these changes preserve the initial meaning and appear perceptually identical to humans, they cause the detector to misclassify the hate speech as non-hate. Word-level attacks modify text by inserting, replacing, or removing words. In Table 1, replacing the word "bitches" with "knuckleheads" causes the model to incorrectly predict the new text as non-hate. Sentence-level attacks directly rewrite the sentence. In Table 1, rephrasing the text "I want to kill you" with "I would love to send you to heaven today" misleads the detector into predicting the text as non-hate. Attacks also occur at multiple levels (i.e., more than one of the above three levels). Prior studies have examined the impact of each type of attack, as shown in Table 2.

Table 2 shows that all four attack types considerably reduce model performance, commonly evaluated by Attack Success Rate (ASR) or after-attack classification metrics. ASR is defined as the proportion of adversarially crafted samples that are misclassified by the target model (Wang et al. 2022). For instance, in Ocampo et al. (2023), 90.7% of the crafted samples can successfully mislead the detector (HateBERT), and hence the ASR is 90.7%. After-attack classification metrics are computed on a combined dataset comprising both adversarially crafted samples (from initially correctly classified samples) and initial clean samples that were misclassified. Based on this dataset, classification metrics such as accuracy, precision, recall, and F1-score are calculated to evaluate the model's performance under adversarial conditions (Huang et al. 2022). For instance, Aggarwal and Vishwakarma (2024) used a character-level attack to generate adversarial samples and combined them with initially misclassified ones to create a new dataset. On this combined dataset, the ALBERT-based hate speech classifier achieved an after-attack accuracy of only 13.1%.

Table 1. Examples of Character-, Word-, and Sentence-Level Adversarial Attacks

| Types | Clean Sample | Adversarial Sample |
|---|---|---|
| Character-level | The South is full of white tra**s**h. The Midwest is full of white tr**a**sh. The West Coast is full of white tras**h**. Predicted label: hate | The South is full of white tra**s**h. The Midwest is full of white tr**a**sh. The West Coast is full of white tras**h**. Predicted label: non-hate |
| Word-level | Derrick said if I cover my face and leave just my eyes visible I'll have tons of **bitches**. Predicted label: hate | Derrick said if I cover my face and leave just my eyes visible I'll have tons of **knuckleheads**. Predicted label: non-hate |
| Sentence-level | **I want to kill you** Predicted label: hate | **I would love to send you to heaven today** Predicted label: non-hate |

Note: character- and word-level examples are from (Aggarwal and Vishwakarma 2024); sentence-level example is from (Azumah et al. 2024).

Table 2. Recent Research on Adversarial Attacks in Harmful Online Content Detection

| Types | Year | Authors | Harmful Content | Detection Models | Attack Methods | Impact of Attacks |
|---|---|---|---|---|---|---|
| Character-level | 2024 | Aggarwal and Vishwakarma | Hate Speech | DistilBERT, LSTM, etc. | Visually confusable glyphs, Zero-width characters | Accuracy: 91.8%→13.1% on ALBERT |
| | 2022 | Li and Chai | Spam | CNN, GRU, LR, NB, etc. | DeepWordBug | F1-score: 82%→44% on GRU |
| | 2019 | Oak and Berkeley | Hate Speech | RF, GB | Word splitting, Word merging, Drop-a-Letter | F1-score: 75%→56% on RF |
| Word-level | 2024 | Zhou et al. | Machine-Generated | RoBERTa | Synonym replacement | AUC: 99.63%→51.06% on RoBERTa |
| | 2023 | Herel et al. | Hate Speech | RoBERTa | TextFooler, TFAdjusted | ASR: 68.3% on RoBERTa |
| | 2023 | K. C. Chen et al. | Misinformation | LSTM, XLNet, RoBERTa, etc. | Synonym replacement, Swap noun and adjective | Accuracy: 77.8%→22.7% on XLNet |
| Sentence-level | 2024 | Przybyła | Misinformation | LSTM, BERT, Gemma | Rephrasing generated by LLM with prompts | ASR: 15.7% on BERT |
| | 2023 | Ocampo et al. | Hate Speech | HateBERT | Few-shot prompting using LLMs | ASR: 90.7% on HateBERT |
| Multi-level | 2025 | Kumbam et al. | Hate Speech | BERT, LSTM, CNN | Explainability-driven synonym replacement and character-level edit | Accuracy after attack/accuracy before attack: 0.7 on LSTM |
| | 2023 | Chang et al. | Hate Speech, Spam | LSTM, CNN, RCNN | Synonym replacement, Dictionary paraphrasing | Accuracy: 99.1%→32.7% on LSTM |

Given the severe ramifications of adversarial attacks, enhancing robustness is essential. As noted earlier, such enhancement has two requirements: first, detectors should handle a wide range of known and new attacks to ensure strong worst-case performance (high generalizability); second, since real-world attacks vary widely, detectors must also maintain strong average performance (high accuracy). Next, we review the robustness enhancement methods to examine whether they satisfy these two requirements.

**2.2 Adversarial Robustness Enhancement Methods**

Various methods have been proposed to enhance adversarial robustness in fields such as natural language processing (NLP) and computer vision. Since some methods developed in other fields may also be used to enhance harmful content detectors, this section reviews methods from both NLP and other fields.

Existing enhancement methods fall into two categories: passive and active (Qiu et al. 2022). Passive methods detect and correct adversarial inputs without modifying ML models. Due to structural differences in data types, these methods are often field-specific. Thus, we focus on the text-based methods, commonly including spelling correction (Pruthi et al. 2019) and abnormal text detection (Mozes et al. 2021). Passive methods work well on character-level attacks. For example, in the aforementioned example that changes "fu**ing immigrants" to "fuking immgrants," spelling correction can effectively work. They can also defend against some word-level attacks, such as those replacing words with rare synonyms. However, they are less effective against more complex attacks, including word-level attacks that use common synonyms or sentence-level attacks. Meanwhile, since they rely on certain patterns, attackers can bypass them by simply avoiding those patterns. Therefore, passive methods are limited in both generalizability and accuracy.

Active methods improve ML models and are generally more effective than the passive ones, making them the focus of most studies (Qiu et al. 2022). Improvement can be approached from three aspects. First, model structure design, which defines the hypothesis space (the set of functions the model can represent, (Anguita et al. 2011)), should favor robust over vulnerable model instances. Second, the parameter learning aspect focuses on how to select, within the hypothesis space, the model instance with the highest adversarial robustness. Third, the model coordination aspect focuses on how to combine the learned model instances, typically by ensemble learning, to enhance robustness. Based on these aspects, existing methods can be grouped into structure-based, learning-based, and ensemble model-based ones, as summarized in Table 3.

**Table 3. Recent Research on Active Methods for Enhancing Adversarial Robustness**

| Types | | | Year | Authors | Methods | Enhanced Models | Data |
|---|---|---|---|---|---|---|---|
| Structure-based | | | 2022 | Nguyen and Tuan | Insert an InfoGAN structure to project adversarial samples onto the embedding space learned from clean samples. | BERT, RoBERTa, etc. | Text |
| | | | 2022 | Zhang et al. | Insert a layer between BERT's output and the classifier to preserve robust features while filtering non-robust ones. | BERT | Text |
| | | | 2021 | S. Zhang et al. | Design a residual-block network to learn precise mappings from adversarial images into clean images. | ResNet, Inception | Image |
| Learning-based | Regularization-based | | 2024 | Liu et al. | Apply gradient norm regularization to reduce prediction variation from adversarial attacks. | HyperIQA, DBCNN, etc. | Image |
| | | | 2022 | Dong et al. | Use a KL divergence to align the model's outputs for clean and adversarial samples. | CNN, LSTM | Text |
| | | | 2022 | Yang et al. | Use triplet loss to bring words closer to their synonyms and push them away from non-synonyms in embedding space. | CNN, LSTM | Text |

| | | | | | | |
|---|---|---|---|---|---|---|
| | Random noise-based | 2024 | X. Zhang et al. | Model word deletion/ insertion/ substitution/ reordering as probability distributions and sample from distributions. | LSTM, BERT | Text |
| | | 2023 | L. Li et al. | Randomly mask words in the input text and fill in the masked words using BERT as a denoising step. | RoBERTa, BERT | Text |
| | | 2022 | Huang et al. | Randomly perturb input based on word frequency for perturbations disruption and distribution alignment. | LSTM, BERT | Text |
| | | 2020 | Ye et al. | Create a synonym set for each word and replace the word by sampling from its synonym set. | CNN, BERT | Text |
| | Adversarial training-based | 2025 | Yang et al. | Fast adversarial training in embedding space by single-step gradient ascent and historical perturbation initialization. | RoBERTa, BERT | Text |
| | | 2024 | Hu et al. | Compute the median of historical model weights during adversarial training to obtain model weights. | WRN | Image |
| | | 2022 | Chen and Ji | Guide the model to treat each word and its substitution in clean and adversarial sample pairs as equally important. | CNN, LSTM, BERT, etc. | Text |
| | | 2021 | Zhao et al. | Combine clean-adversarial representations to create mixup samples, and align their predictive distributions. | CNN, LSTM, BERT | Text |
| Ensemble model-based | | 2024 | Waghela et al. | Dynamic ensemble with a meta-model to weight base models for ensemble predictions. | BERT, ALBERT, etc. | Text |
| | | 2023 | Qin et al. | Dynamic ensemble selection: choose the base model prediction with the lowest uncertainty. | WRN | Image |
| | | 2022 | Li and Chai | Train base models on varied samples, enhance robustness via adversarial training, and ensemble predictions. | LSTM, GRU, LR, etc. | Text |
| | | 2020 | Sen et al. | Combine a full-precision (32-bit) model with a low-precision (4- or 2-bit) one. | CNN, AlexNet, etc. | Image |

**2.2.1 Structure-Based Adversarial Robustness Enhancement Methods**

Structure-based methods refine the hypothesis space by modifying the model's structure, such as adding an embedding projection module (Nguyen and Tuan 2022) or a robust feature extraction module (Zhang et al. 2022). For instance, Nguyen and Tuan (2022) add an InfoGAN-based module to project adversarial samples onto the embedding space of the initial clean samples. However, these methods often require careful structural changes, limiting their compatibility since existing ML models must be redesigned. Meanwhile, complex ML models have broad hypothesis spaces (Leshno et al. 1993). Even if the initial model instance is not robust, the space may still contain many robust alternatives. Conversely, even if the space is refined to include a higher proportion of robust instances, the selected one may still have low robustness. The critical factor, therefore, lies in the selection process, i.e., how the parameters are learned.

**2.2.2 Learning-Based Adversarial Robustness Enhancement Methods**

These methods improve the training process to increase the chance of selecting a robust model from the hypothesis space, including regularization-, random noise-, and adversarial training-based methods.

**1) Regularization-Based Methods**

Regularization-based methods add regularization terms to the training objective to bias learning toward robust models. Since text is discrete, regularizations in other fields, such as gradient regularization (Liu et al. 2024) and Lipschitz regularization (Fazlyab et al. 2023) for images, are less effective for textual data. Common text-based regularizations include word embedding regularization, which encourages synonyms to have similar embeddings (Yang et al. 2022), and output-level regularization, which aligns the predictive distribution of adversarial samples with those of their initial clean samples (Dong et al. 2021). However, regularizations are typically attack-specific and lack generalizability. For example, the word embedding regularization targets word-level attacks like synonym substitutions but fails against character- or sentence-level attacks, while the output-level regularization relies on specific adversarial samples, limiting its effectiveness on new attacks. Besides, choosing the right regularization strength is challenging: too much may harm accuracy on clean samples, while too little provides limited benefits (Waseda et al. 2025). Thus, regularization-based methods may struggle to achieve high generalizability and accuracy.

**2) Random Noise-Based Methods**

Random noise-based methods introduce random perturbations (noise) to input samples during training. For images, this can involve adding Gaussian or Laplacian noise (Lecuyer et al. 2019). For text, words can be randomly replaced with synonyms (Ye et al. 2020, Zhou et al. 2021) or masked and then reconstructed using a language model (Li et al. 2023), generating multiple input variants. By enforcing consistent predictions between the variants and the original input, these methods encourage the model to learn representations robust to noise, including adversarial perturbations. A key limitation is that such noise is easy to apply at the character or word level but difficult at the sentence level, reducing generalizability.

These methods also struggle to meet the accuracy requirement. During testing, random noise generates multiple input variants, thus producing multiple predictions. The final prediction is obtained using statistical or aggregation methods. Statistical methods assume predictions follow a specific distribution (e.g., binomial distribution for binary classification (Zeng et al. 2023)), and compute the lower confidence bound of the predictions. If the bound exceeds a threshold (e.g., 0.5 for binary classification (Zeng et al. 2023)), the predictions are considered certified. Certified radius, i.e., the maximum allowable perturbation to the input, can also be calculated based on the predictions. This approach can obtain the final prediction with certified robustness, guaranteeing that the model remains robust under perturbations within a certain radius. However, its accuracy is limited due to two reasons. First, it assumes that all constituent parts of an input (e.g., words) are perturbed with equal probability. However, adversarial attacks tend to target the most sensitive parts (e.g., words with greater influence on the prediction (Jin et al. 2020)), reducing both the practical utility of the computed radius and the defense accuracy. Second, this approach assumes all predictions are identically distributed, neglecting that some may be more accurate and thus should carry more weight. Unlike statistical methods, aggregation methods focus on weighting predictions to aggregate them. For example, in Zhou et al. (2021), predictions are aggregated using a CBW-D algorithm, which gives higher weights to predictions with larger margins (the gap between the highest and other probabilities), addressing the second limitation of statistical methods. But the accuracy of this approach is still limited. First, like the statistical approach, it also fails to consider the more sensitive parts when generating variants. Second, it considers prediction differences only during testing, not training, reducing its effectiveness.

### 3) Adversarial Training-Based Methods

Adversarial training (AT) was first proposed in the image field by Goodfellow et al. (2015). It emulates attackers by using existing attack methods to generate adversarial samples, and an ML model is then trained on these samples with the objective of producing correct predictions for adversarial inputs. AT is also popular in NLP. For instance, Zhao et al. (2021a) proposed mixup regularized adversarial training (MRAT), which first uses attack methods (e.g., DeepWordBug (Gao et al. 2018)) to generate adversarial samples and then optimizes the model using a mixup-based strategy on paired clean and adversarial samples. AT addresses the limitation of random noise-based methods by focusing on the most sensitive parts through prioritizing the most threatening variants, namely adversarial samples. Learning directly from these samples significantly enhances adversarial robustness. However, AT has a key limitation: it requires prior knowledge of attacks to generate adversarial samples. As a result, its robustness is mostly confined to attacks seen during training and remains vulnerable to new ones, thereby limiting its generalizability. Meanwhile, AT alone is insufficient and is often combined with ensemble learning to enhance robustness.

However, the current combinations are still limited and can reduce accuracy (illustrated next).

**2.2.3 Ensemble Model-Based Adversarial Robustness Enhancement Methods**

Ensemble model-based methods combine multiple learned ML models (called *base models*) and aggregate their predictions. In this case, adversarial samples must deceive a majority of the base models, which is substantially more difficult than fooling a single model. This increased difficulty raises the barrier for attackers, thereby enhancing the overall adversarial robustness (Li and Chai 2022). These methods have been widely applied in fields such as images and text. For image, Sen et al. (2020) combined a full-precision (32-bit) model with a low-precision (4- or 2-bit) model, leveraging the former's strength on clean samples and the latter's on adversarial ones. For text, Li and Chai (2022) combined multiple base models such as CNN, LSTM, and GRU, aggregating their predictions to detect adversarial spam emails and fake reviews.

However, current methods fall short of the generalizability requirement, as advanced transferable attacks can deceive multiple base models simultaneously. For instance, Jin et al. (2020) demonstrated that adversarial samples carefully crafted for a CNN-based text classifier not only achieved a 100% ASR on the CNN but also transferred to LSTM- and BERT-based classifiers with ASRs of 84.9% and 90.2%. This reduces the robustness of ensemble models against such attacks. Meanwhile, they fail to meet the accuracy requirement. The ensemble's robustness depends on its base models, which are often vulnerable, limiting overall robustness. Some studies combine ensemble learning with AT, applying it to each base model to improve their robustness (Li and Chai 2022, Waghela et al. 2024). However, the enhancement remains limited because these studies primarily apply AT to strengthen the base models, while neglecting the weight assignment component (i.e., the weight assignor). Since the weight assignor directly determines each model's influence on the ensemble's final prediction, neglecting its robustness may result in assigning higher weights to less robust base models, thereby reducing the ensemble's overall adversarial robustness.

## 3. An Overview of Our Proposed Method

Enhancing adversarial robustness requires both generalizability and accuracy. Drawing on transfer learning (Arjovsky et al. 2019), where generalizability is measured by the worst-case performance across domains, we extend this idea to our study. Let $\mathcal{A}$ denote the set of all textual adversarial attacks. We define the generalizability of a detector $D$ as its worst-case performance: $\text{Gen}(D) \coloneqq \min_{a \in \mathcal{A}} P(D, a)$, where $P(D, a)$ denotes the performance of detector $D$ under attack $a$. We define the accuracy as the expected performance: $\text{Acc}(D) \coloneqq \mathbb{E}_{a \in \mathcal{A}} P(D, a)$. We aim to maximize both $\text{Gen}(D)$ and $\text{Acc}(D)$ in this study.

The key difference between our study and prior studies is that prior studies mainly maximize the expected performance over a subset of attacks, i.e., $\mathbb{E}_{a \in \mathcal{A}'} P(D, a)$, where $\mathcal{A}' \subset \mathcal{A}$ (e.g., character-level attacks in spelling error correction (Pruthi et al. 2019) or known attacks in AT (Goodfellow et al. 2015)). Prior studies typically overlook $\text{Gen}(D)$. As a result, $\text{Gen}(D)$ decreases, meaning the detector may perform well on some attacks but fail on others. This flaw poses a critical risk because it allows attackers to craft targeted attacks that exploit the detector's specific weaknesses, an issue that is especially pronounced given the inherently adversarial nature of such attacks. Meanwhile, $\text{Acc}(D)$ also declines, not only because the detector is less effective to defend against attacks outside $\mathcal{A}'$, but also because its ability to handle attacks within $\mathcal{A}'$ is constrained by the limitations we discussed in the literature review. This is also critical because $\text{Acc}(D)$ reflects the detector's overall performance. Over time, a detector may face a diverse array of attacks initiated by distinct adversaries, and a lower $\text{Acc}(D)$ elevates the risk of frequent failures across them.

However, directly maximizing $\text{Gen}(D)$ and $\text{Acc}(D)$ is intractable. A detector $D$ is a discrete ML model rather than a continuous variable as in classical optimization, resulting in a discrete search space (i.e., the set of all possible model configurations) that is hard to optimize. To address this, we follow the computational design science paradigm, developing a new IT artifact that addresses current limitations and performs the focal task more effectively. Instead of computing $\text{Gen}(D)$ and $\text{Acc}(D)$ and then optimizing over $D$, we design a new $D$ to address the limitations and empirically evaluate its improvements on real data. Since the ideal detector corresponds to optimal performance, observed improvements indicate the proposed $D$ is closer to the ideal and solves the optimization problem more effectively than prior studies.

Even following the computational design science paradigm, two challenges remain. First, a vast number of adversarial attacks have been devised, and new ones continue to emerge. Moreover, attacks are discrete in nature and difficult to characterize mathematically. Consequently, the set $\mathcal{A}$ is vast, potentially infinite, and intractable. Designing a detector to achieve both objectives across such a complex set is therefore highly challenging. Second, jointly achieving the two objectives is itself difficult. Traditional ML focuses on maximizing expected accuracy by optimizing a loss defined on a set of training samples, aiming to obtain a model that performs well on general cases. Consequently, the model may overfit to common attacks: improving $\text{Acc}(D)$ by boosting performance on frequent attacks while degrading performance on rarer attacks, thereby reducing $\text{Gen}(D)$. Although techniques such as transfer learning and multi-task learning can leverage knowledge across domains or coordinate learning across tasks, they are not suitable for our setting. They assume that the objectives are computable so that coordination can be performed. For example, in multi-task learning, a common approach involves computing losses for samples across different tasks, summing these losses, calculating the gradients of the sum with respect to model parameters, and updating the parameters accordingly (Zhang and Yang 2018). In our case, however, $\text{Acc}(D)$ and $\text{Gen}(D)$ are defined over an intractable attack set $\mathcal{A}$, rendering both their values and their gradients hard to compute. To address these two challenges, we adopt a sequential approach. We first propose a framework ensuring that any detector instantiated within it can defend against a broad range of attacks, thereby securing high generalizability. We then propose a specific instantiation within this framework that focuses on improving accuracy. In this way, our instantiated detector achieves high generalizability and high accuracy.

## 4. An LLM-based Sample Generation and Aggregation Framework (LLM-SGA)

### 4.1 Core Idea of Our LLM-SGA Framework

Textual adversarial attacks occur at the character, word, and sentence levels, with new attacks continually emerging. This challenges defenses, which must counter a variety of known and new attacks. Our core idea is to identify the key invariances of all attacks and base our framework on these invariances. Before introducing our framework, we define the concept of the identical meaning set.

**Definition (Identical Meaning Set)**: It is defined as the collection of all texts that convey the same underlying meaning. Formally, $\mathcal{S}(h) = \{x | x \text{ is a text AND } s(x) = h\}$, where $s(x)$ denotes the underlying meaning of text $x$ and $\mathcal{S}(h)$ denotes the set of all texts whose meaning is $h$. For example, the two texts "I want to kill you" and "I would love to send you to heaven today" in Table 1 convey the same underlying meaning (i.e., $h$ represents "a threat to kill"), and thus they belong to the same identical meaning set.

Let an initial clean sample be $x^C$ with ground truth $y^C$. An adversarial sample ($x^A$) preserves the meaning of $x^C$ but causes the detector to make an incorrect prediction, whereas it predicts $x^C$ correctly. Hence, an adversarial sample $x^A$ is defined by two properties. First, $x^A$ and $x^C$ belong to the same identical meaning set *(Invariance 1)*. This is essential because if the meaning is altered, even if it evades the detection, the functional intent of the attack is lost. For instance, if the text "I want to kill you" is changed to "I want to play basketball," the new text may evade detection, but it no longer conveys the attacker's malicious intent, rendering the attack ineffective. Hence, an attacker will avoid such modifications. The second invariance is that attacks typically arise when the targeted detector performs well, particularly when it correctly predicts the initial clean sample $x^C$ *(Invariance 2)*. Only in such cases do attackers need to craft adversarial samples to evade detection; otherwise, $x^C$ already fulfills its intended function.

From Invariance 2, the detector is expected to correctly predict samples that share the same meaning as the clean input. However, this is not always the case, as adversarial samples can still fool the detector. Our LLM-SGA enhances robustness by making predictions over multiple samples rather than a single one, thereby diluting the impact of any individual sample, including a potential adversarial one. This mechanism is effective against all adversarial samples, ensuring strong worst-case performance for high generalizability.

In practice, the input sample may be adversarial or clean, but it is unclear to the detector. When the input is adversarial ($x^A$), it implies that its corresponding clean sample $x^C$ has been detected. But the detector does not know $x^C$. However, since $x^A$ and $x^C$ share the same meaning, we can generate multiple samples (denoted as $\hat{x}_1, \hat{x}_2, \ldots, \hat{x}_N$) that preserve the meaning of $x^A$, and therefore also that of $x^C$ (as per Invariance 1). When the input is clean ($x^C$), we can also generate multiple samples preserving the same meaning as $x^C$. For simplicity, we denote the input as $x_0$, which can be $x^A$ or $x^C$. We generate samples that preserve the meaning of $x_0$, and make a detection by aggregating the predictions for them. Since $x_0$ can be adversarial or clean, and adversarial samples can include character-, word-, or sentence-level perturbations, generating such samples is nontrivial. To address this, we use LLMs for generation. Since many advanced

LLMs are restricted from responding to potentially harmful content, we adopt the open-source LLaMA.

**4.2 Description of Our LLM-SGA Framework**

Our LLM-SGA framework includes three steps.

**Step 1:** Given an input $x_0$, we prompt the LLaMA to generate samples $(\hat{x}_1, \hat{x}_2, ..., \hat{x}_N)$:

"*The following text may contain adversarial perturbations generated by techniques such as word misspelling, synonym substitution, and sentence rephrasing. Please generate N new texts by rephrasing the input text while preserving its original meaning.*
*INPUT {text}*
*OUTPUT:*"

**Step 2:** We fed $x_0$ and the generated $\hat{x}_1, \hat{x}_2, ..., \hat{x}_N$ into the detector to obtain predictive probabilities $p_0, ..., p_N \in [0,1]$. Each value represents the likelihood that the label is 1 (harmful) according to the detector.

**Step 3:** We aggregate the predictive probabilities by computing their average ($\bar{p}$):

$$\bar{p} = \frac{1}{N+1} \sum_{n=0}^{N} p_n \quad (1)$$

We then compare $\bar{p}$ with a threshold $\varepsilon$ to get the predicted label $\hat{y}$ to decide whether the input $x_0$ is harmful:

$$\hat{y} = \begin{cases} 0 \text{ (i.e., non-harmful), if } \bar{p} \leq \varepsilon \\ 1 \text{ (i.e., harmful), if } \bar{p} > \varepsilon \end{cases} \quad (2)$$

**4.3 Theoretical Properties of Our LLM-SGA Framework**

Let $p$ denote the predictive probability of sample $x_0$, and $y$ its ground truth label. The expectation and variance of $p$ are denoted as: $\mu_0 = \mathbb{E}[p]$ and $\sigma_0^2 = \text{VAR}[p]$. Since $\hat{x}_1, \hat{x}_2, ..., \hat{x}_N$ and $x_0$ belong to the same identical meaning set, each corresponding $p_n$ has the same expectation $\mu_0$ and variance $\sigma_0^2$. Thus, the $\bar{p}$ has expectation $\mu_0$ and variance $\sigma_0^2/(N+1)$. Let $P(\hat{y} = y)$ denote the probability that the $\hat{y}$ is correct.

We consider a basic random guessing case where the detector's output is uniformly distributed over [0,1]. If the value exceeds $\varepsilon$, the predicted label is 1; otherwise, it is 0. Assuming the detector performs better than random guessing (a reasonable assumption given its overall excellent performance), we define $\delta = |\mu_0 - \varepsilon|$. We then have the following inequality (see Appendix B for details):

$$P(\hat{y} = y) \geq 1 - \sigma_0^2/[(N+1)\delta^2] \quad (3)$$

Hence, the probability of correct detection is lower bounded. Moreover, increasing the number of samples

$N$ raises this lower bound, indicating that generating and aggregating more samples is advantageous.

## 5. Instantiation: Adversarially Robust Harmful Online Content Detector (ARHOCD)

### 5.1 Overview of Our Instantiated Detector

The LLM-SGA ensures that a detector instantiated within it is generalizable. But the accuracy still depends on a specific instantiation. As mentioned earlier, accurate detection faces three key challenges, which we address with three corresponding novel design components. Briefly, first, to tackle the single-detector challenge where relying on a single detector is highly vulnerable, we propose a two-dimensional ensemble that integrates multiple base detectors (Design Component 1). Second, to address the weighting challenge where more accurate predictions should receive higher weights, we propose a Bayesian two-dimensional weight assignment method that combines prior knowledge with actual data to obtain a neural network-based weight assignor that determines weight for each prediction in the matrix (Design Component 2). Third, to overcome the learning challenge where the parameters of both the base detectors and the weight assignor should be jointly made adversarially robust, we introduce an iterative adversarial training strategy that iteratively optimizes the base detectors and the weight assignor (Design Component 3). Together, these three design components constitute our instantiated detector (ARHOCD) as illustrated in Figure 1.

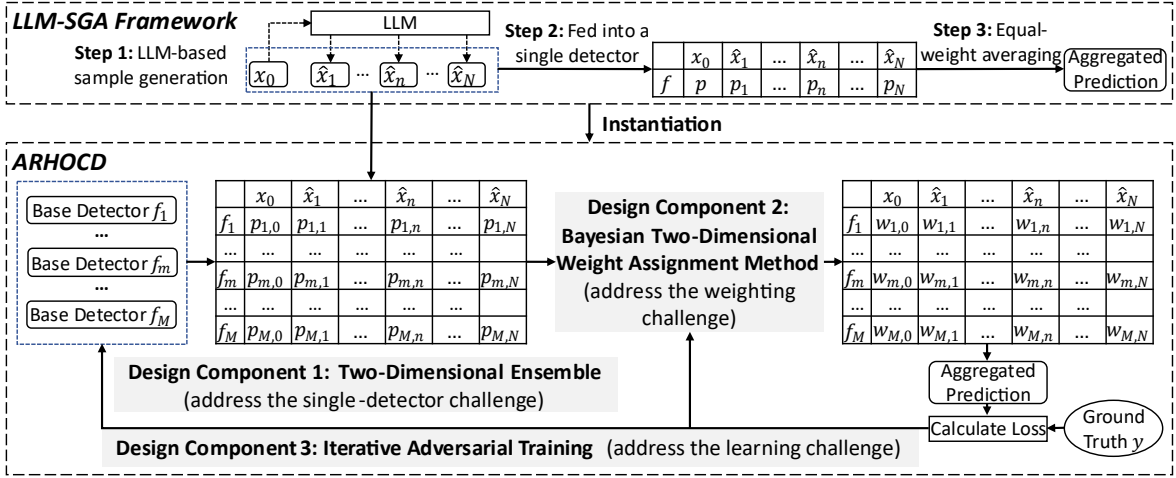

Figure 1. Overview of Our ARHOCD

### 5.2 Design Component 1: A Two-Dimensional Ensemble

We propose to ensemble $M$ existing detectors. Each detector, such as a BERT-based model, is referred to as a base detector to distinguish it from the overall instantiated detector, ARHOCD, which is composed of multiple base detectors. We denote the $m$-th base detector as $f_m$, whose predictive probability for the input sample $x_0$ is $p_{m,0}$, and for a generated sample $\hat{x}_n$ is $p_{m,n}$. The predictive probabilities of $f_m$ form a vector $\boldsymbol{p}_m = [p_{m,0}, p_{m,1}, \ldots, p_{m,n}, \ldots, p_{m,N}]$. The predictive probabilities of all base detectors form a two-dimensional matrix $\boldsymbol{P} = [\boldsymbol{p}_1, \ldots, \boldsymbol{p}_m, \ldots, \boldsymbol{p}_M]^\mathrm{T}$. We then aggregate them by average:

$$\bar{\bar{p}} = \frac{1}{M(N+1)} \sum_{m=1}^{M} \sum_{n=0}^{N} p_{m,n} \tag{4}$$

Then, $\bar{\bar{p}}$ is compared with a threshold $\varepsilon$ to determine whether the input $x_0$ is harmful, as in Equation (2).

Introducing multiple base detectors enhances accuracy. We assume that each base detector performs better than random guessing. For samples sharing the same meaning as $x_0$, each base detector's predictive probability is a random variable with expectation $\mu_m$ and variance $\sigma_m^2$. We can prove that (Appendix C):

$$P(\hat{y} = y) \geq 1 - \frac{1}{(N+1)M^2\delta^2} \sum_{m=1}^{M} \sigma_m^2 \tag{5}$$

where $\delta = |(1/M) \sum_{m=1}^{M} \mu_m - \varepsilon|$. As $M$ increases, the numerator $\sum_{m=1}^{M} \sigma_m^2$ grows linearly, while $M^2$ in the denominator grows quadratically and $\delta$ remains relatively stable. As a result, the lower bound of $P(\hat{y} = y)$ increases, indicating that ensembling more base detectors is beneficial, provided that each one performs better than random guessing, which is typically easy to meet. Meanwhile, increasing $N$ also raises the lower bound, suggesting that incorporating more samples is advantageous.

In summary, by ensembling multiple base detectors, we obtain a two-dimensional matrix of predictive probabilities for aggregation. Unlike prior ensemble studies that aggregate either samples or base models into a one-dimensional prediction vector, our study simultaneously incorporates both to form a matrix. We also theoretically demonstrate the advantage of our two-dimensional ensemble structure.

### 5.3 Design Component 2: A Bayesian Two-Dimensional Weight Assignment Method

#### 5.3.1 Overview of the Second Design Component

Consistent with prior studies (Glenn 1950), we use the Brier Score, defined as $(p_{m,n} - y)^2$, to measure the accuracy of each predictive probability $p_{m,n}$. A lower score indicates a prediction closer to the ground truth, reflecting better calibration and accuracy (Glenn 1950). Since different predictive probabilities correspond to different accuracy, a viable approach is weighted aggregation, assigning higher weights to more accurate ones. Hence, Equation (4) becomes:

$$\bar{\bar{p}} = \sum_{m=1}^{M} \sum_{n=0}^{N} w_{m,n} p_{m,n} \qquad (6)$$

$w_{m,n}$ is the weight assigned for $p_{m,n}$, satisfying two constraints: $\sum_{m=1}^{M} \sum_{n=0}^{N} w_{m,n} = 1$ and $w_{m,n} \geq 0, \forall m, n$. We denote the likelihood of making the correct prediction $y$ as $p(y|\boldsymbol{P})$. Since $p(y|\boldsymbol{P})$ depends on the values of $w_{m,n}, \forall m, n$, assigning appropriate weights $w_{m,n}$ for each $p_{m,n}$ is crucial.

Weight assignment methods can be categorized as selection or weighting ones. For selection method, it selects a subset of predictions (each from a base model or for a sample) for the final output (Qin et al. 2023). It eliminates the influence of less accurate predictions by assigning them zero weights. However, enforcing exact zero weights may impair flexibility and compromise performance. Weighting method, by contrast, is more general and flexible, as it is not strictly forced to zero out any predictions. Hence, we focus on this approach, which can be further divided into fixed- and dynamic-weight methods. Fixed-weight methods assign constant weights to each base model during training, which remain fixed at runtime (Li and Chai 2022). However, this is unsuitable in our context because samples $(\hat{x}_1, \hat{x}_2, \dots, \hat{x}_N)$ are generated at runtime and thus their weights cannot be predetermined. Dynamic-weight methods adapt weights based on the given sample (Zhou et al. 2021). While valuable, they still face two limitations. First, they assign weights to a vector of predictions, considering either multiple samples or multiple base models, but not both. In our case, prediction accuracy depends on both the sample and the base model, making weight assignment more complex. Second, these methods usually assume that higher predictive probabilities indicate higher accuracy. This assumption fails in the context of adversarial attacks, where adversarial samples can still yield high predictive probabilities (e.g., an adversarial image was misclassified as a gibbon with 99.3% predictive probability (Goodfellow et al. 2015)). Hence, determining the values of $w_{m,n}$ for each sample and base detector under adversarial conditions remains a challenging task.

We propose a dynamic weight assignment method with three phases. Phase 1 follows prior studies by introducing intermediate weights and log-normal distributions to satisfy the two constraints. Since the values of $w_{m,n}$ are unobserved, we employ Bayesian learning to infer them based on both prior knowledge and observed data. Such integration results in more effective weight assignment. Phase 2 designs the prior. Since the accuracy of each prediction depends jointly on the sample and the base detector, we decompose it into two factors: the sample's predictability and the base detector's capability. We theoretically show that a sample's predictability is negatively correlated with the variance of predictions across base detectors, while a base detector's capability is negatively correlated with the variance of its predictions across samples. These relationships guide us in designing a prior, which serves as the core technical novelty of our Bayesian weight assignment method. Phase 3 infers the posterior weights. As the true posterior $p(w_{m,n}|P, y)$ is intractable, we approximate it using variational inference with a self-attention-based neural network.

### 5.3.2 Phase 1: Introduce Intermediate Weight and Log-Normal Distribution to Satisfy Constraints

We introduce intermediate weights $\widetilde{w}_{m,n}$ to meet the first constraint (i.e., $\sum_{m=1}^{M}\sum_{n=0}^{N} w_{m,n} = 1$):

$$w_{m,n} = \widetilde{w}_{m,n} / \sum_{m=1}^{M}\sum_{n=0}^{N} \widetilde{w}_{m,n} \tag{7}$$

We further require $\widetilde{w}_{m,n} \geq 0$ to meet the second constraint (i.e., $w_{m,n} \geq 0$). Thus, the task becomes how to determine $\widetilde{w}_{m,n}$. Each $\widetilde{w}_{m,n}$ is modeled as a log-normal distribution to ensure its non-negativity. Formally,

$$p(\widetilde{w}_{m,n}) = \text{LogNormal}(\psi_{m,n}, \sigma^2) \tag{8}$$

$\psi_{m,n}$ and $\sigma^2$ are the mean and variance of the Gaussian distribution underlying the log-normal distribution.

### 5.3.3 Phase 2: Design the Prior Distribution

**1) Disentangle Factors of Sample and Base Detector**

The accuracy of $p_{m,n}$ depends on two factors: sample $\hat{x}_n$'s predictability (when $n = 0$, $\hat{x}_n$ refers to $x_0$) and detector $f_m$'s capability. A *sample's predictability* reflects how likely it is to be correctly predicted and is measured by the Brier Score of its predictions across models: $e_n = \frac{1}{M}\sum_{m=1}^{M}(p_{m,n} - y)^2$. A *detector's capability* reflects its overall ability to make correct predictions and is measured by the Brier Score of its predictions across samples: $e_m = \frac{1}{N+1}\sum_{n=0}^{N}(p_{m,n} - y)^2$. Higher weights should be assigned to samples with greater predictability and to base detectors with stronger capability. To capture this, we factorize $w_{m,n}$ as: $w_{m,n} = w_m w_n$, where $w_m$ is the overall weight for detector $f_m$ ($w_m = \sum_n w_{m,n}$) and $w_n$ is the overall weight for sample $\hat{x}_n$ ($w_n = \sum_m w_{m,n}$). We prove in Appendix D that the intermediate weight $\widetilde{w}_{m,n}$ can also be factorized as: $\widetilde{w}_{m,n} = \widetilde{w}_m \widetilde{w}_n$ with $w_m \propto \widetilde{w}_m$ and $w_n \propto \widetilde{w}_n$, where $\propto$ denotes proportionality. Thus, it suffices to determine $\widetilde{w}_m$ and $\widetilde{w}_n$. Since both are non-negative, log-normal distributions are adopted:

$$p(\widetilde{w}_m) = \text{LogNormal}(\psi_m, \sigma_{\text{dt}}^2); \quad p(\widetilde{w}_n) = \text{LogNormal}(\psi_n, \sigma_{\text{sp}}^2) \tag{9}$$

**2) Identify Domain Knowledge**

In matrix $P$, each column represents the predictive probabilities of a sample. For the $n$-th column, the mean and variance are $\bar{p}_n = \frac{1}{M}\sum_{m=1}^{M} p_{m,n}$ and $\sigma_n^2 = \frac{1}{M}\sum_{m=1}^{M}(p_{m,n} - \bar{p}_n)^2$. We have (see Appendix E):

$$\mathbb{E}[e_n] = \mathbb{E}[\sigma_n^2] + \mathbb{E}[(\bar{p}_n - y)^2] \tag{10}$$

Since all the samples have the same meaning, their expected predictive probabilities are identical, i.e., $\mathbb{E}[\bar{p}_n]$ is the same for all samples. With the same $y$, $\mathbb{E}[(\bar{p}_n - y)^2]$ is also identical across samples. Hence, a larger $\mathbb{E}[\sigma_n^2]$ implies a larger $\mathbb{E}[e_n]$. This makes the observed $\sigma_n^2$ a useful indicator of sample $\hat{x}_n$'s lower predictability. Hence, a larger $\sigma_n^2$ should correspond to a smaller $w_n$ and thus a smaller $\widetilde{w}_n$. Since a smaller $\psi_n$ in the log-normal distribution leads to a smaller expected $\widetilde{w}_n$, $\psi_n$ should also be smaller.

A similar conclusion holds for detector $f_m$, whose predictions correspond to the $m$-th row of matrix $P$, with mean $\bar{p}_m = \frac{1}{N+1}\sum_{n=0}^{N} p_{m,n}$ and variance $\sigma_m^2 = \frac{1}{N+1}\sum_{n=0}^{N}(p_{m,n} - \bar{p}_m)^2$. Then, $\mathbb{E}[e_m] = \mathbb{E}[\sigma_m^2] + \mathbb{E}[(\bar{p}_m - y)^2]$ (see Appendix E). In our ensemble, base detectors perform comparably across a wide range of samples. While predictive probabilities for individual samples may vary across detectors, these variations tend to cancel out when averaged over many samples. Thus, we assume $\mathbb{E}[\bar{p}_m]$ is identical across detectors, which is also empirically validated in Appendix F. Estimating $\mathbb{E}[\sigma_m^2]$ with the observed $\sigma_m^2$, a larger $\sigma_m^2$ implies a larger $\mathbb{E}[e_m]$, which corresponds to a smaller $w_m$ (and thus $\widetilde{w}_m$), and consequently, a smaller $\psi_m$.

**3) Design Prior Distribution based on Disentangled Factors and Domain Knowledge**

Since $\sigma_n^2$ and $\sigma_m^2$ are inversely related to $\psi_n$ and $\psi_m$, we define $\psi_n$ and $\psi_m$ as,

$$\psi_n = \exp[-\alpha \cdot \sigma_n^2]; \quad \psi_m = \exp[-\beta \cdot \sigma_m^2] \tag{11}$$

$\alpha$ and $\beta$ are positive hyperparameters. The exponential function helps strictly suppress the impact of highly inaccurate predictions (e.g., such as adversarial ones that can greatly increase variance). Since we factorize $\widetilde{w}_{m,n}$ into two factors, $\widetilde{w}_{m,n} = \widetilde{w}_m \widetilde{w}_n$, we show in Appendix G that for the $p(\widetilde{w}_{m,n})$ in Equation (8), its $\psi_{m,n}$ and $\sigma^2$ can be expressed as $\psi_{m,n} = \psi_m + \psi_n$, $\sigma^2 = \sigma_{dt}^2 + \sigma_{sp}^2$. Then, we have:

$$p(\widetilde{w}_{m,n}) = \text{LogNormal}\left(\exp[-\alpha \cdot \sigma_n^2] + \exp[-\beta \cdot \sigma_m^2], \sigma_{dt}^2 + \sigma_{sp}^2\right) \quad (12)$$

In Equation (12), we actually design a mapping function from $\boldsymbol{P}$ (from which $\sigma_n^2$ and $\sigma_m^2$ are computed) to $\psi_{m,n}$ and then get $p(\widetilde{w}_{m,n})$. Technically, it should be denoted as $p(\widetilde{w}_{m,n}|\boldsymbol{P})$, but we still use $p(\widetilde{w}_{m,n})$ for simplicity. $p(\widetilde{w}_{m,n})$ serves as a prior for two reasons. First, it embeds useful domain knowledge (assigning smaller weights to higher-variance predictions), but the expectations are estimated from a single instance (e.g., using $\sigma_m^2$ to estimate $\mathbb{E}[\sigma_m^2]$), introducing randomness that may weaken the domain knowledge. Second, while valuable, the multiplicative disentanglement and exponential functional form may not be optimal and oversimplify the underlying relationship. Thus, $p(\widetilde{w}_{m,n})$ is useful but still needs refinement. Hence, we use it as a prior to guide the learning of assigning more appropriate weights, i.e., the posterior.

**5.3.4 Phase 3: Obtain the Posterior**

The posterior is advantageous as the mapping function from $\boldsymbol{P}$ is refined with the observed ground truth $y$, resulting in more appropriate weight assignments. We denote the posterior of $\widetilde{w}_{m,n}$ as $p(\widetilde{w}_{m,n}|\boldsymbol{P}, y)$, modeled as $\text{LogNormal}(\varphi_{m,n}^{\text{post}}, (\sigma^{\text{post}})^2)$. The task is to get the parameters $\varphi_{m,n}^{\text{post}}$ and $\sigma^{\text{post}}$. However, their exact values are intractable. Hence, we adopt the neural network-based variational inference (Kingma and Welling 2014), using a neural network (*inference network*) to output the parameters of the variational distribution (i.e., $\varphi_{m,n}^{\text{post}}$ and $\sigma^{\text{post}}$). To simplify inference and reduce complexity, we focus on inferring $\varphi_{m,n}^{\text{post}}$ and set $(\sigma^{\text{post}})^2$ directly as $\sigma_{dt}^2 + \sigma_{sp}^2$ (same as the prior). Since the ground truth $y$ is unobserved at runtime, the inference network cannot rely on it as input and should instead take only $\boldsymbol{P}$ as input. We denote the inference network for obtaining $\varphi_{m,n}^{\text{post}}$ as $\varphi_{m,n}^{\text{post}} = f^{\text{infer}}(\boldsymbol{P})$, which is trained to make the variational distribution approximate the true posterior $p(\widetilde{w}_{m,n}|\boldsymbol{P}, y)$. Thus, the task reduces to learning $f^{\text{infer}}(\boldsymbol{P})$.

Our inference network is built on self-attention. Since each row of $\boldsymbol{P}$ corresponds to a base detector, and each column corresponds to a sample, we first apply self-attention to the rows and columns to capture dependencies within each detector and sample, and then combine them to model their interactions. Let $\boldsymbol{p}_m$ denote the predictions of detector $f_m$ and $\boldsymbol{p}_n$ those of sample $\hat{x}_n$. We use self-attention to get representation $z_m$ for $p_m$ and $z_n$ for $p_n$, and then combine them to predict $\varphi_{m,n}^{\text{post}}$ with a multilayer perceptron (MLP):

$$z_m = \text{SelfAttention}(p_m); \quad z_n = \text{SelfAttention}(p_n); \quad \varphi_{m,n}^{\text{post}} = \text{MLP}(z_m; z_n) \qquad (13)$$

The structural details of the SelfAttention(·) and MLP(·) are in Appendix H. Compared with the prior that only uses variance, self-attention allows us to capture more intricate and useful dependencies. Moreover, the factors of the sample and the base detector are combined through an MLP instead of a specific multiplicative form, allowing to model richer interactions. In addition, a neural network is used as the inference network instead of a specific exponential function, enabling more effective extraction of information from $P$ to determine weights. Let $\widetilde{W}$ collectively denote $\widetilde{w}_{m,n}$ and $W$ denote $w_{m,n}$. Let $\phi$ denote the parameters of the inference network, which outputs $\varphi_{m,n}^{\text{post}}$ to define the variational distribution $q_\phi(\widetilde{W})$ over weights $\widetilde{W}$ (and thus over $W$). Hence, the inference network serves as the *weight assignor*.

Since we approximate the posterior $p(\widetilde{W}|P, y)$ with a variational distribution $q_\phi(\widetilde{W})$, the log-likelihood $\log p(y|P)$ computed with $q_\phi(\widetilde{W})$ is a lower bound of the true $\log p(y|P)$ (see Appendix I):

$$\mathcal{L} = \mathbb{E}_{q_\phi(\widetilde{W})}[\log p(y|P, \widetilde{W})] - \sum_{m=1}^{M} \sum_{n=0}^{N} \text{KL}\left(q_\phi(\widetilde{w}_{m,n}) \| p(\widetilde{w}_{m,n})\right) \qquad (14)$$

where $\mathcal{L}$ denotes the lower bound. The first term of the lower bound represents the model's likelihood of making a correct prediction based on $\widetilde{W}$ from the variational distribution. The second term measures the Kullback-Leibler divergence between the variational and the prior distribution. Their detailed computations are in Appendix J. We learn $\phi$ by maximizing $\mathcal{L}$. In this way, the weight assignor is learned from both the observed data ($P$ and $y$) and the identified domain knowledge that has been embedded in our prior.

### 5.4 Design Component 3: An Iterative Adversarial Training (IAT) Strategy

Adversarial training (AT) has been shown to learn models with robust parameters, and we adopt this idea in our study. However, unlike the prior AT studies that focus only on enhancing base models, we use AT to enhance both the base detectors and the weight assignor. This introduces a key challenge: the base detectors and the weight assignor jointly determine the final prediction, yet the dataset for training base detectors cannot be reused for training the weight assignor, which must evaluate the base detectors on a separate dataset to assign weights accordingly. Reusing the same data would cause overfitting and biased weight assignment. To address this issue, we propose an iterative adversarial training (IAT) strategy.

We denote the dataset to train base detectors and the weight assignor as $\mathcal{D}^{bd}$ and $\mathcal{D}^{ag}$, respectively. As with prior work, we use existing attack methods to craft adversarial samples for each sample in $\mathcal{D}^{bd}$ and $\mathcal{D}^{ag}$, yielding datasets $\widetilde{\mathcal{D}}^{bd}$ and $\widetilde{\mathcal{D}}^{ag}$. They are then combined with the initial datasets to get two augmented datasets: $\mathcal{D}^{bd} \cup \widetilde{\mathcal{D}}^{bd}$ and $\mathcal{D}^{ag} \cup \widetilde{\mathcal{D}}^{ag}$. Detailed process is in Appendix K. For each $(x_i^{bd}, y_i^{bd}) \in \mathcal{D}^{bd} \cup \widetilde{\mathcal{D}}^{bd}$, we generate $N$ samples with the same meaning as $x_i^{bd}$ using LLM, denoted as $\hat{x}_{i,1}^{bd}, \ldots, \hat{x}_{i,N}^{bd}$. The initial $x_i^{bd}$ and the generated $\hat{x}_{i,1}^{bd}, \ldots, \hat{x}_{i,N}^{bd}$ are collectively denoted as $\boldsymbol{x}_i^{bd}$. The predictive probabilities for $\boldsymbol{x}_i^{bd}$ form a two-dimensional matrix $\boldsymbol{P}_i^{bd}$. Its weight matrix is denoted as $\boldsymbol{W}_i^{bd}$. The final prediction is computed as $\mathrm{tr}\left[(\boldsymbol{W}_i^{bd})^T \boldsymbol{P}_i^{bd}\right]$. We define the loss over $\mathcal{D}^{bd} \cup \widetilde{\mathcal{D}}^{bd}$ as the cross-entropy (CE) loss. Then, we have:

$$\mathrm{Loss}^{bd} = \sum_{i=1}^{|\mathcal{D}^{bd} \cup \widetilde{\mathcal{D}}^{bd}|} \mathrm{CE}\left(\mathrm{tr}\left[(\boldsymbol{W}_i^{bd})^T \boldsymbol{P}_i^{bd}\right], y_i^{bd}\right) \tag{15}$$

The $m$-th row of $\boldsymbol{P}_i^{bd}$ is produced by the detector parameterized by $\boldsymbol{\theta}_m$, while $\boldsymbol{W}_i^{bd}$ comes from the weight assignor parameterized by $\boldsymbol{\phi}$. Since $\mathcal{D}^{bd} \cup \widetilde{\mathcal{D}}^{bd}$ is used to train the base detectors, $\mathrm{Loss}^{bd}$ updates only $\boldsymbol{\theta}_m$ ($\forall m$) with $\boldsymbol{\phi}$ fixed. We update $\boldsymbol{\theta}_m$ using the Adam optimizer with an initial learning rate $\eta$:

$$\boldsymbol{\theta}_m \leftarrow \boldsymbol{\theta}_m - \mathrm{AdamUpdate}(\partial \mathrm{Loss}^{bd}/\partial \boldsymbol{\theta}_m, \eta) \tag{16}$$

Using similar notations, we define the objective over $\mathcal{D}^{ag} \cup \widetilde{\mathcal{D}}^{ag}$ as maximizing the sum of the lower bounds of each sample it contains (as shown in Equation 14). The corresponding loss is then defined as:

$$\mathrm{Loss}^{ag} = \sum_{i=1}^{|\mathcal{D}^{ag} \cup \widetilde{\mathcal{D}}^{ag}|} \left[-\mathbb{E}_{q_\phi(\widetilde{\boldsymbol{W}}_i^{ag})}[\log p(y_i^{ag}|\boldsymbol{P}_i^{ag}, \widetilde{\boldsymbol{W}}_i^{ag})] + \mathrm{KL}\left(q_\phi(\widetilde{\boldsymbol{W}}_i^{ag}) \| p(\widetilde{\boldsymbol{W}}_i^{ag})\right)\right] \tag{17}$$

Since $\mathcal{D}^{ag} \cup \widetilde{\mathcal{D}}^{ag}$ is used to train the weight assignor, $\mathrm{Loss}^{ag}$ is used to update only $\boldsymbol{\phi}$ with $\boldsymbol{\theta}_m$ ($\forall m$) fixed:

$$\boldsymbol{\phi} \leftarrow \boldsymbol{\phi} - \mathrm{AdamUpdate}(\partial \mathrm{Loss}^{ag}/\partial \boldsymbol{\phi}, \eta) \tag{18}$$

We alternate between Equations (16) and (18) until convergence (pseudocode in Appendix L), jointly learning the parameters of base detectors and the weight assignor for enhanced robustness. A technical challenge is that $\boldsymbol{W}_i^{ag}$ is a random sample from the distribution $q_\phi$, making gradient computation and the update process difficult. We address this using the reparameterization trick, with details in Appendix M.

While some studies combine AT with ensemble learning, they apply AT only to the base detectors, neglecting the weight assignor and its interactions with base detectors. In contrast, our IAT robustifies both the base detectors and the weight assignor and captures their interactions. Notably, in practice, base detectors may already be deployed, so practitioners may keep them fixed to ensure higher compatibility or reduce training costs. This consideration is particularly relevant for base detectors with a large number of parameters, such as transformer-based models, where retraining can be costly. In such case, our IAT can be simplified: only using Equation (18) to train the weight assignor, without updating the base detectors.

### 5.5 Predicting New Samples

After training, we obtain our ARHOCD. For a new input $x^{\text{new}}$, its prediction involves three steps. First, we prompt the LLM to generate $N$ samples: $\hat{x}_1^{\text{new}}, \ldots, \hat{x}_n^{\text{new}}, \ldots, \hat{x}_N^{\text{new}}$. Second, we apply its base detectors to all samples to obtain a two-dimensional matrix of predictive probabilities $\boldsymbol{P}^{\text{new}}$, with each element denoted as $p_{m,n}^{\text{new}}$. Third, we apply its weight assignor to get the weight $w_{m,n}^{\text{new}}$ for $p_{m,n}^{\text{new}}$. As $w_{m,n}^{\text{new}}$ is a random variable, we compute the expectation. Then, the predictive probability $\hat{p}^{\text{new}}$ is given by,

$$\hat{p}^{\text{new}} = \sum_{m=1}^{M} \sum_{n=0}^{N} \left( \mathbb{E}[w_{m,n}^{\text{new}}] \cdot p_{m,n}^{\text{new}} \right) \tag{19}$$

If $\hat{p}^{\text{new}} \leq \varepsilon$, the input $x^{\text{new}}$ is classified as non-harmful; otherwise, it is classified as harmful. However, $\mathbb{E}[w_{m,n}^{\text{new}}]$ is intractable. We propose two solutions: using Monte Carlo (MC) method to approximate $\mathbb{E}[w_{m,n}^{\text{new}}]$, or computing a lower bound of $\mathbb{E}[w_{m,n}^{\text{new}}]$ to represent it. The corresponding ARHOCDs are denoted as Ours (MC) and Ours (LB), respectively. The details of these two solutions are in Appendix N.

## 6. Evaluation

### 6.1 Experiment Settings

#### 6.1.1 Datasets

We evaluated our method on three widely used datasets: ETHOS, PHEME and White Supremacist (De Gibert et al. 2018, Mollas et al. 2022). ETHOS is a dataset for hate speech detection. It contains tweets from Hatebusters and Reddit, each annotated by five Figure-Eight crowdworkers to indicate if the tweet contains hate speech. PHEME is a rumor detection dataset with tweets on news events, labeled by professional journalists. White Supremacist is an extremist content detection dataset, with posts from the Stormfront forum, each labeled by three experts. Table 4 shows their statistics (details in Appendix O). To handle class imbalance in the PHEME and White Supremacist datasets, we used the cost-sensitive learning strategy proposed in (Ting 2002) during training, where higher weights are assigned to the minority class.

Table 4. Statistics of the Datasets

| Datasets | #Samples | Class Ratio (Positive/Negative) | Dataset Split (Train/Val/Test) | Avg. Len. | Max. Len. | Min. Len. |
|---|---|---|---|---|---|---|
| ETHOS | 998 | Hate (44.39%)/ Non-hate (55.61%) | 798/100/100 | 23.24 | 606 | 2 |
| PHEME | 6425 | Rumor (37.39%)/ Non-rumor (62.61%) | 5140/642/643 | 21.46 | 38 | 3 |
| White Supremacist | 10,319 | Hate (11.59%)/ Non-hate (88.41%) | 8255/1032/1032 | 18.68 | 366 | 1 |

**6.1.2 The Assumption of Attacker and Defender's Knowledge in the Experiments**

We assume attackers know the details of the detector being used, which is a common assumption in prior studies (Yoo and Qi 2021). This assumption is grounded in two reasons. First, many detectors' structures or source codes are often publicly accessible, and even if not, attackers could infer or obtain them through various means, such as inference or network intrusion. Second, following Kerckhoffs's principle, which states that a system should remain secure even if all its information is public, adversarial robustness should be evaluated and enhanced assuming the detector's details are known to attackers.

Defenders, however, are often unaware of which attack will be launched. Since the attack set is vast, potentially infinite, it is infeasible to enumerate every possible attack. Based on the literature, we selected four representative attacks covering four levels to simulate adversaries. For character-level attack, we chose DeepWordBug (Gao et al. 2018, Li and Chai 2022), which generates adversarial samples by performing four character-level perturbations: substitution, insertion, deletion, and neighboring swap. For word-level attack, we chose TFAdjusted (Herel et al. 2023, Morris et al. 2020), which deletes each word and measures the change in the model's prediction to identify key words, and then replaces the key words with synonyms. For sentence-level attack, we chose TREPAT (Przybyła et al. 2025), which splits long texts into smaller segments, rephrases these segments with an LLM using diverse prompts, extracts edits (e.g., substitutions, insertions, and deletions), and then applies these edits via beam search to generate adversarial samples. For multi-level attack, we chose Explainability-Driven adversarial attack (ExplainDrive, (Kumbam et al. 2025)). It first identifies the most important features using interpretability techniques (e.g., LIME, (Ribeiro et al.

2016)), which assign importance scores based on each feature's contribution to the prediction. It then adversarially perturbs the most important features at different levels to generate adversarial samples.

### 6.2 Experiment 1: Generalizability Comparison

#### 6.2.1 Baselines and Metrics

We evaluated the generalizability of our LLM-SGA by showing that a detector instantiated within it outperforms detectors robustified by baselines. Since encoder-only transformers excel in classification tasks, we used BERT (the most popular model in this family) as the detector. The baselines cover the major types of adversarial robustness enhancement methods reviewed earlier, whose core ideas are summarized in Table 5 and details in Appendix P. For passive defenses, we combined BERT with LLM-puri and Det&Res. For active defenses, structure-based methods were excluded because they modify detector structures, whereas ours preserves them for better compatibility. We included learning-based methods where BERT was fine-tuned. We also included ensemble model-based methods. Since these methods use multiple base detectors, we included BERT and its four variants (DistilBERT, RoBERTa, XLNet, ALBERT) as base detectors. For fair comparison, we instantiated two detectors within LLM-SGA: one using only BERT (LLM-SGA (single)) and one using the same five base detectors as the ensemble baselines (LLM-SGA (multiple)).

Table 5. Brief Descriptions of Baselines

| Baselines | | | | Brief Descriptions |
|---|---|---|---|---|
| Passive | LLM-puri (Moraffah et al. 2024) | | | Prompt an LLM to purify adversarial samples. |
| | Detection and Restoration (Det&Res) (Wang et al. 2022) | | | Detect and restore adversarial samples. |
| Active | Learning-based | Random noise-based | SAFER (Ye et al. 2020) | Generate multiple samples by synonym replacement. |
| | | | MASKFil (Li et al. 2023) | Generate multiple samples by masked word filling. |
| | | Regularization-based | FTML (Yang et al. 2022) | Bring words closer to synonyms and push them away from non-synonyms in the embedding space. |
| | | | OutReg (Dong et al. 2021) | Align outputs for clean and adversarial sample pairs. |
| | | | FIM (Gloeckler et al. 2023) | Penalize gradient directions with high sensitivity. |
| | | Adversarial training-based | AT (Goodfellow et al. 2015) | Incorporate adversarial samples into the training set. |
| | | | MRAT (Zhao et al. 2021) | Optimize the model using a mixup-based strategy. |
| | | | FAT (Yang et al. 2025) | Fast adversarial training in the embedding space. |
| | Ensemble model-based | ARText (Li and Chai 2022) | | Use adversarial training and model ensemble. |
| | | ARDEL (Waghela et al. 2024) | | Use a meta-model to weight base models. |
| | | EnsSel (Qin et al. 2023) | | Select predictions with the lowest uncertainties. |

We used the four aforementioned attacks to generate adversarial samples from test set for evaluation. For each detector, we computed ASR and after-attack classification metrics (described in Section 2.1, with computation details in Appendix Q). The worst-case performance across attacks reflects generalizability.

### 6.2.2 Results

We repeated the experiment 20 times, reporting the mean and standard deviation (std) and examining statistical significance via the t-test. To save space, we only report the ASRs of the four attacks on the PHEME dataset in Table 6, while other metrics and the results of the other two datasets are reported in Appendix R.1. For simplicity, a robustified detector is denoted by the enhancement method applied to it. For instance, the detector robustified by Det&Res is denoted as "Det&Res" in Table 6.

Table 6. Comparison with Baselines on Generalizability (ASR%)

| Detectors | DeepWordBug | TFAdjusted | TREPAT | ExplainDrive | Worst-Case |
|---|---|---|---|---|---|
| Raw BERT | 56.86*** (2.81) | 58.28*** (2.53) | 43.60*** (3.89) | 74.95*** (3.04) | 74.95 |
| LLM-puri | 19.42** (3.77) | 20.98*** (1.91) | 19.80** (3.32) | 34.54* (2.09) | 34.54 |
| Det&Res | 31.79*** (2.28) | 30.24*** (3.03) | 33.54*** (3.91) | 41.18*** (4.75) | 41.18 |
| SAFER | 47.32*** (3.32) | 47.20*** (2.63) | 45.43*** (4.40) | 61.43*** (3.89) | 61.43 |
| MASKFil | 52.40*** (4.20) | 63.49*** (2.60) | 43.65*** (2.72) | 73.81*** (4.08) | 73.81 |
| FTML | 43.85*** (4.26) | 43.84*** (2.81) | 47.25*** (3.76) | 64.53*** (4.70) | 64.53 |
| OutReg | 44.53*** (2.71) | 41.96*** (3.45) | 37.64*** (4.25) | 56.54*** (5.01) | 56.54 |
| FIM | 56.53*** (2.81) | 59.05*** (3.62) | 52.26*** (4.16) | 72.85*** (5.03) | 72.85 |
| AT | 50.64*** (3.33) | 48.70*** (2.94) | 39.67*** (3.37) | 63.41*** (3.36) | 63.41 |
| MRAT | 46.04*** (2.88) | 53.18*** (3.00) | 46.52*** (2.58) | 72.95*** (2.88) | 72.95 |
| FAT | 56.32*** (3.33) | 57.07*** (4.17) | 43.63*** (4.66) | 72.12*** (3.92) | 72.12 |
| LLM-SGA (single) | **16.02 (2.70)** | **16.36 (2.38)** | **23.34 (3.55)** | **31.91 (4.59)** | **31.91** |
| ARText | 24.23*** (3.57) | 21.35*** (3.20) | 23.59*** (4.03) | 26.97*** (3.66) | 26.97 |
| ARDEL | 21.04*** (3.33) | 18.75*** (3.27) | 20.18*** (3.48) | 26.95*** (5.35) | 26.95 |
| EnsSel | 17.07*** (2.92) | 15.96 (2.51) | 24.43*** (3.58) | 21.86* (2.72) | 24.43 |
| LLM-SGA (multiple) | **12.08 (2.47)** | **15.22 (1.99)** | **14.95 (2.91)** | **19.68 (3.74)** | **19.68** |

Note: A higher ASR indicates that more adversarial samples bypass the detector, resulting in lower performance; *$p < 0.05$; **$p < 0.01$; ***$p < 0.001$; Values in brackets show std. Bold font marks the best results. Same below.

By comparing our LLM-SGA (single) with the baselines, we find that the BERT detector robustified by LLM-SGA is more generalizable. Specifically, it obtains the best worst-case performance across all attacks, with an ASR of 31.91%. This means that, in the worst case, when attackers generate 100 adversarial samples, only about 32 of them can fool the detector. In contrast, among the baselines, the detector robustified by LLM-puri performs best, with its worst-case results under the ExplainDrive attack. In this setting, it attains an ASR of 34.54%, significantly higher than that of ours ($p < 0.05$). Meanwhile, by comparing LLM-SGA (multiple) with ensemble model-based baselines, we find that ensembling multiple base detectors and robustifying them with LLM-SGA yields the best performance, with a worst-case ASR of 19.68%. The baselines ARText, ARDEL, EnsSel obtain worst-case ASRs of 26.97%, 26.95% and 24.43%, respectively—all significantly higher than that of ours ($p < 0.001$). By the way, comparing LLM-SGA (multiple) with LLM-SGA (single) shows that ensembling multiple base detectors increases detection accuracy, reducing the ASR across the four attacks from 16.02%, 16.36%, 23.34%, and 31.91% to 12.08%, 15.22%, 14.95%, and 19.68%. This demonstrates the effectiveness of our first design component.

### 6.3 Experiment 2: Accuracy Comparison

We compared our ARHOCD with detectors robustified by baselines to show its advantage in accuracy. ARHOCD includes five base detectors (BERT, DistilBERT, RoBERTa, XLNet, and ALBERT). Since ARHOCD includes the processes of generating samples, aggregating predictions from the generated samples and base detectors, and training parameters, while some baselines only include one or two processes, we enhanced all baselines to include comparable processes. This ensures that any performance gain arises from our novel design components rather than from simply assembling more processes.

The enhancement principle is as follows: if a baseline already contains a given process, we retain its own version; otherwise, we enhance it with a well-performing one. For example, the regularization-based methods (FTML, OutReg, FIM) do not generate samples, so we incorporated SAFER (a well-performing sample generation method according to our experiments). Similarly, because these baselines use a single detector, we extended them to five base detectors as ours and then aggregated their outputs using CBW-D (a well-performing aggregation method according to our experiments). However, since these baselines have their own training processes, the detector parameters were learned using their respective own ones. With the same principle, we enhanced all the other baselines (details in Appendix S), as summarized in Table 7. Since all detectors used the same five base detectors, we omit this detail from Table 7 for brevity.

Experiments were conducted in two scenarios: defending known attacks and new attacks. For known attacks, each dataset was split into training ($D^{\text{train}}$) and test ($D^{\text{test}}$) sets. Adversarial samples were generated for each set by randomly applying one of three attacks: DeepWordBug, TFAdjusted, or TREPAT, yielding adversarial datasets: $D_{\text{DTT}}^{\text{train}}$ (from $D^{\text{train}}$) and $D_{\text{DTT}}^{\text{test}}$ (from $D^{\text{test}}$). Detectors were robustified (e.g., adversarially trained) on $D_{\text{DTT}}^{\text{train}}$ and evaluated on $D_{\text{DTT}}^{\text{test}}$. For new attacks, an adversarial test set $D_{\text{E}}^{\text{test}}$ was crafted from $D^{\text{test}}$ using the ExplainDrive attack, which was unseen in the robustification process and thus regarded new. All methods were evaluated on $D_{\text{E}}^{\text{test}}$. Since baselines FTML, SAFER, MASKFil, and FAT do not use adversarial samples during robustification, all attacks are new for them. Hence, these baselines were included only in new-attack scenario. For a fair comparison, they were also evaluated on $D_\text{E}^\text{test}$.

Table 7. Enhancements of Baselines in Experiment 2

| Types | | Enhanced Baselines | Samples | Aggregation | Learning |
|---|---|---|---|---|---|
| Passive defense baselines | | LLM-puri | Own | Lack (CBW-D) | AT |
| | | Det&Res | Own | Lack (CBW-D) | AT |
| Active defense baselines | Random noise-based | SAFER | Own | Lack (CBW-D) | Own |
| | | MASKFil | Own | Lack (CBW-D) | Own |
| | Regularization-based | FTML | Lack (SAFER) | Lack (CBW-D) | Own |
| | | OutReg | Lack (SAFER) | Lack (CBW-D) | Own |
| | | FIM | Lack (SAFER) | Lack (CBW-D) | Own |
| | Adversarial training-based | AT | Lack (SAFER) | Lack (CBW-D) | Own |
| | | MRAT | Lack (SAFER) | Lack (CBW-D) | Own |
| | | FAT | Lack (SAFER) | Lack (CBW-D) | Own |
| | Ensemble model-based | ARText | Lack (SAFER) | Own | Own |
| | | ARDEL | Own | Own | Own |
| | | EnsSel | Lack (SAFER) | Own | Own |

For both scenarios, we conducted an ablation study to examine the contribution of each key design component in our ARHOCD. First, to examine the two-dimensional ensemble (Design Component 1), we removed the model ensemble, retaining only a single detector (Ours w/o ME). Second, we tested the weight assignment method (Design Component 2). We removed our variance-based prior (Ours w/o prior) and the data-driven refinement process that aims to approximate the posterior (Ours w/o data). We replaced our weight assignment method with a fixed-weight baseline (Ours w/ FEL), a dynamic-weight baseline using CBW-D (Ours w/ CBW-D), and a selection method that, for each sample, chooses the detector with the highest predictive probability (Ours w/ ES). While not explored in prior work, we also devised a probability-based prior that used the gap between the predictive probabilities of the two classes: a larger gap indicates higher reliability and thus gets a larger weight. We replaced our variance-based prior with it, yielding Ours w/ Prob-prior. Third, we tested our IAT (Design Component 3) by removing it (Ours w/o IAT). In addition, since current AT enhances only the base detectors, we also evaluated a variant where IAT was replaced with AT applied only to the base detectors (Ours w/ AT). Appendix T shows further details of these variants.

For both scenarios, we also investigated the sensitivity of key parameters in our detector, including the number of generated samples, the number of base detectors, and the $\sigma^2$ of the log-normal distribution.

In summary, Experiment 2 comprises three sub-experiments: (1) baseline comparison, (2) ablation analysis, and (3) sensitivity analysis, each performed under two scenarios on three datasets. For brevity, we present results on the PHEME dataset in the manuscript, with the other two datasets in Appendix R.2.

**6.3.1 Results of Experiment 2.1 (Comparison with Baselines in Defending Known and New Attacks)**

Table 8 reports the experimental results. Ours (MC) and Ours (LB), which use Monte Carlo and lower-bound approximations of Equation (19), respectively, perform comparably and outperform all enhanced baselines, showing strong robustness in harmful content detection against known and new attacks. For instance, for known attacks, Ours (LB) improves after-attack accuracy, precision, recall, and F1-score over the best baseline (LLM-puri (Enhanced)) by 2.3%, 2.0%, 4.2%, and 3.0% ($p < 0.001$). For new attacks, Ours (MC) reduces ASR from 15.38% of the best baseline (MRAT (Enhanced)) to 13.16% ($p < 0.001$).

**Table 8. Results of Comparison with Baselines in Defending Known and New Attacks**

| Attack Scenario | Method | Performance after attack↑ (%) | | | | ASR↓ (%) |
|---|---|---|---|---|---|---|
| | | Accuracy | Precision | Recall | F1-score | |
| Known Attacks (DeepWordBug, TFAdjusted, TREPAT) | LLM-puri (Enhanced) | 83.66*** (0.67) | 82.73*** (0.71) | 82.09*** (0.80) | 82.38*** (0.75) | 12.55*** (0.67) |
| | Det&Res (Enhanced) | 81.38*** (0.77) | 80.42*** (0.81) | 79.27*** (0.95) | 79.74*** (0.89) | 14.99*** (1.12) |
| | OutReg (Enhanced) | 81.15*** (0.62) | 80.60*** (0.63) | 78.36*** (0.80) | 79.16*** (0.75) | 15.26*** (1.22) |
| | AT (Enhanced) | 81.59*** (0.85) | 80.82*** (0.89) | 79.23*** (1.04) | 79.85*** (0.99) | 15.03*** (1.14) |
| | MRAT (Enhanced) | 83.54*** (1.06) | 82.74*** (1.21) | 81.71*** (1.09) | 82.15*** (1.13) | 13.96*** (0.96) |
| |ARText (Enhanced) | 80.95*** (0.75) | 79.62*** (0.80) | 79.82*** (0.78) | 79.71*** (0.78) | 12.79*** (0.81) |
| | ARDEL (Enhanced) | 80.78*** (0.96) | 80.10*** (1.08) | 78.08*** (1.11) | 78.81*** (1.10) | 15.03*** (1.14) |
| | EnsSel (Enhanced) | 81.81*** (0.84) | 81.32*** (0.97) | 79.13*** (0.92) | 79.93*** (0.93) | 15.35*** (0.74) |
| | Ours (MC) | **85.59 (0.76)** | **84.42 (0.78)** | **85.54 (0.86)** | **84.85 (0.80)** | 10.59 (1.04) |
| | Ours (LB) | 85.49 (0.82) | 84.32 (0.85) | 85.45 (0.93) | 84.76 (0.87) | **10.59 (1.03)** |
| New Attacks (ExplainDrive) | LLM-puri (Enhanced) | 82.41 (0.92) | 81.41 (1.03) | 80.65*** (0.98) | 81.02*** (1.01) | 14.24** (0.55) |
| | Det&Res (Enhanced) | 80.05*** (0.94) | 78.93*** (1.01) | 77.89*** (1.08) | 78.32*** (1.05) | 16.27*** (1.40) |
| | SAFER (Enhanced) | 77.27*** (1.11) | 75.78*** (1.22) | 75.20*** (1.17) | 75.45*** (1.18) | 17.54*** (1.13) |
| | MASKFil (Enhanced) | 77.15*** (1.23) | 75.77*** (1.38) | 74.61*** (1.33) | 75.06*** (1.34) | 18.69*** (1.04) |
| | FTML (Enhanced) | 78.13*** (0.76) | 76.98*** (0.88) | 75.40*** (0.82) | 75.99*** (0.83) | 19.20*** (1.36) |
| | OutReg (Enhanced) | 79.88*** (0.77) | 79.36*** (0.85) | 76.71*** (0.91) | 77.59*** (0.90) | 17.52*** (1.38) |
| | FIM (Enhanced) | 73.51*** (0.97) | 73.05*** (1.17) | 68.21*** (1.21) | 68.98*** (1.32) | 25.36*** (1.56) |
| | AT (Enhanced) | 80.57*** (0.81) | 79.63*** (0.87) | 78.19*** (0.96) | 78.76*** (0.92) | 16.29*** (0.88) |
| | MRAT (Enhanced) | 82.05** (0.99) | 81.22 (1.17) | 79.91*** (1.05) | 80.44*** (1.07) | 15.38*** (1.20) |
| | FAT (Enhanced) | 73.59*** (0.94) | 73.32*** (1.19) | 68.17*** (1.15) | 68.94*** (1.27) | 26.42*** (1.06) |
| | ARText (Enhanced) | 79.16*** (1.23) | 77.74*** (1.30) | 77.82*** (1.35) | 77.77*** (1.32) | 15.11*** (1.27) |
| | ARDEL (Enhanced) | 79.04*** (0.99) | 78.29*** (1.13) | 75.94*** (1.16) | 76.73*** (1.15) | 16.29*** (0.88) |
| | EnsSel (Enhanced) | 79.90*** (0.94) | 79.68*** (1.05) | 76.42*** (1.11) | 77.43*** (1.11) | 18.44*** (1.28) |
| | Ours (MC) | **82.91 (0.66)** | **81.67 (0.67)** | **82.61 (0.63)** | **82.03 (0.66)** | **13.16 (1.12)** |
| | Ours (LB) | 82.88 (0.66) | 81.64 (0.68) | 82.58 (0.58) | 82.00 (0.65) | 13.24 (1.13) |

**6.3.2 Results of Experiment 2.2 (Ablation Analysis in Defending Known and New Attacks)**

Table 9 presents the results, revealing three observations. First, using a single detector without model ensemble greatly degrades performance. For instance, defending known attacks with only BERT lowers the F1-score from 84.85% to 75.99%, showing lower robustness. Second, removing the variance-based prior (Ours w/o prior) or replacing it (Ours w/ Prob-prior) degrades performance. Removing the data-driven refinement process (Ours w/o data) also worsens performance. Meanwhile, alternative weighting methods, including Ours w/ FEL, Ours w/ CBW-D and Ours w/ ES, perform worse. For instance, in defending new attacks, Ours w/ CBW-D rises ASR from 13.16% to 15.26%. Finally, removing or replacing IAT (Ours w/o IAT, Ours w/ AT) notably reduces robustness. For instance, removing IAT reduces after-attack accuracy from 85.59% to 81.70% for known attacks and from 82.91% to 79.29% for new attacks. Overall, removing or replacing any key design component degrades performance, highlighting the contribution of each.

Table 9. Results of Ablation Analysis in Defending Known and New Attacks

| Attack Scenario | Method | | Performance after attack↑ (%) | | | | ASR↓ (%) |
|---|---|---|---|---|---|---|---|
| | | | Accuracy | Precision | Recall | F1-score | |
| **Known Attacks** (DeepWordBug, TFAdjusted, TREPAT) | Ours w/o ME | BERT | 77.26*** (0.45) | 75.73*** (0.47) | 76.35*** (0.50) | 75.99*** (0.48) | 14.88*** (0.53) |
| | | DistilBERT | 75.23*** (0.75) | 73.74*** (0.78) | 74.64*** (0.81) | 74.04*** (0.79) | 18.08*** (0.85) |
| | | RoBERTa | 78.94*** (0.65) | 77.58*** (0.66) | 78.67*** (0.69) | 77.95*** (0.67) | 14.78*** (0.70) |
| | | XLNet | 75.81*** (0.56) | 74.64*** (0.59) | 75.97*** (0.65) | 74.93*** (0.60) | 13.56*** (0.65) |
| | | ALBERT | 77.32*** (0.72) | 75.85*** (0.81) | 74.98*** (0.72) | 75.35*** (0.75) | 17.88*** (0.64) |
| | Ours w/o prior | | 84.77** (0.87) | 83.57* (0.91) | 84.36*** (0.93) | 83.91** (0.91) | 11.24* (0.69) |
| | Ours w/o data | | 85.03* (0.59) | 83.90 (0.64) | 84.31*** (0.64) | 84.09** (0.62) | 11.29* (0.54) |
| | Ours w/ FEL | | 84.55*** (0.55) | 83.40*** (0.60) | 83.75*** (0.55) | 83.56*** (0.56) | 11.83*** (0.50) |
| | Ours w/ CBW-D | | 84.91* (0.69) | 83.90 (0.75) | 83.80*** (0.75) | 83.84*** (0.74) | 11.67*** (0.65) |
| | Ours w/ ES | | 84.98* (0.75) | 84.06 (0.83) | 83.65*** (0.78) | 83.84*** (0.79) | 12.07*** (0.98) |
| | Ours w/ Prob-prior | | 84.75** (0.66) | 83.58** (0.65) | 84.41*** (0.85) | 83.90*** (0.71) | 11.63** (0.85) |
| | Ours w/o IAT | | 81.70*** (0.80) | 80.61*** (0.88) | 79.96*** (0.84) | 80.25*** (0.85) | 13.64*** (1.10) |
| | Ours w/ AT | | 84.23*** (0.54) | 83.07*** (0.56) | 83.52*** (0.62) | 83.27*** (0.58) | 11.25* (0.68) |
| | Ours (MC) | | **85.59 (0.76)** | **84.42 (0.78)** | **85.54 (0.86)** | **84.85 (0.80)** | 10.59 (1.04) |
| | Ours (LB) | | 85.49 (0.82) | 84.32 (0.85) | 85.45 (0.93) | 84.76 (0.87) | **10.59 (1.03)** |
| **New Attacks** (ExplainDrive) | Ours w/o ME | BERT | 65.08*** (0.59) | 63.21*** (0.58) | 63.63*** (0.60) | 63.33*** (0.59) | 30.67*** (0.76) |
| | | DistilBERT | 63.85*** (0.78) | 62.76*** (0.79) | 63.50*** (0.84) | 62.71*** (0.80) | 31.80*** (0.89) |
| | | RoBERTa | 69.23*** (0.50) | 67.86*** (0.54) | 68.74*** (0.59) | 68.03*** (0.54) | 26.17*** (0.60) |
| | | XLNet | 68.77*** (0.79) | 67.40*** (0.81) | 68.26*** (0.86) | 67.56*** (0.82) | 23.05*** (0.74) |
| | | ALBERT | 66.42*** (0.90) | 63.96*** (0.96) | 63.72*** (0.93) | 63.82*** (0.94) | 30.51*** (0.81) |
| | Ours w/o prior | | 81.95** (1.01) | 80.67** (1.06) | 81.05** (1.15) | 80.84*** (1.09) | 14.21* (1.30) |
| | Ours w/o data | | 82.38* (0.77) | 81.20 (0.80) | 81.10*** (0.94) | 81.14** (0.87) | 14.04* (0.86) |
| | Ours w/ FEL | | 81.69*** (0.68) | 80.49*** (0.72) | 80.22*** (0.80) | 80.35*** (0.75) | 14.80*** (0.84) |
| | Ours w/ CBW-D | | 81.91*** (0.94) | 80.86** (1.06) | 80.15*** (0.98) | 80.46*** (1.00) | 15.26*** (1.68) |
| | Ours w/ ES | | 81.84*** (0.95) | 80.84** (1.06) | 79.92*** (1.02) | 80.31*** (1.03) | 16.48*** (1.01) |
| | Ours w/ Prob-prior | | 82.26** (0.65) | 81.07** (0.60) | 81.43** (1.34) | 81.16** (0.92) | 14.57** (1.30) |
| | Ours w/o IAT | | 79.29*** (1.00) | 78.04*** (1.07) | 77.19*** (1.16) | 77.55*** (1.12) | 16.80*** (1.32) |
| | Ours w/ AT | | 82.47 (0.64) | 81.23 (0.68) | 81.51*** (0.71) | 81.36** (0.69) | 13.89* (0.85) |
| | Ours (MC) | | **82.91 (0.66)** | **81.67 (0.67)** | **82.61 (0.63)** | **82.03 (0.66)** | **13.16 (1.12)** |
| | Ours (LB) | | 82.88 (0.66) | 81.64 (0.68) | 82.58 (0.58) | 82.00 (0.65) | 13.24 (1.13) |

**6.3.3 Results of Experiment 2.3 (Sensitivity Analysis in Defending Known and New Attacks)**

Figure 2 shows results for defending known (a-1 to a-3) and new (b-1 to b-3) attacks. First, increasing the number of generated samples $N$ improves after-attack performance (accuracy, recall, precision, F1) and reduces ASR, aligning with our theoretical analysis that more samples enhance performance. Second, more base detectors $M$ also boost performance, consistent with our theoretical analysis that ensembling more base detectors enhances detection accuracy. Third, the variance term $\sigma^2$ of the log-normal distribution does not significantly affect performance, validating our focus on the mean term (i.e., $\psi$) rather than on $\sigma^2$.

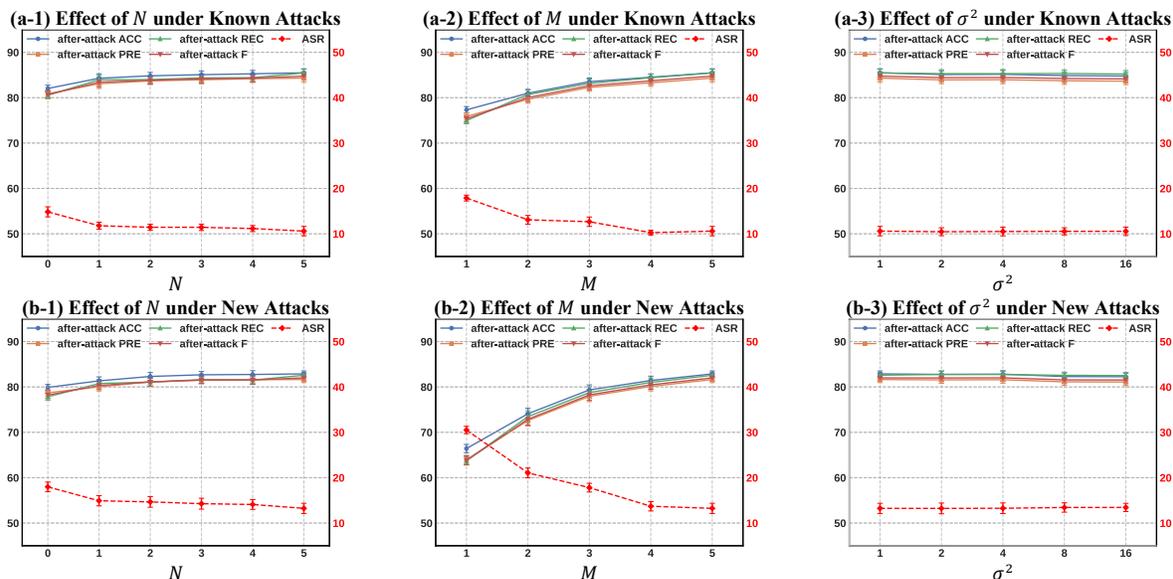

**Figure 2. Results of Sensitivity Analysis in Defending Known and New Attacks**

### 6.4 Experiment 3: Accuracy Comparison without Updating Detectors

As noted earlier, defenders may choose to keep the base detector parameters fixed. In this case, only the weight assignor is updated. Under this setting, we compared our detector with the baselines (Experiment 3.1), conducted ablation analysis (Experiment 3.2) and sensitivity analysis (Experiment 3.3) across the three datasets. Results are in Appendix U. Conclusions align with those from Experiment 2. Meanwhile, we find that fixing the base detectors leads to inferior performance than updating them, which is expected because updating enhances base detectors and thereby enhances the robustness of our ARHOCD. Whether to update the base detectors depends on practitioners' judgment: if the attacks are severe enough, they should update base detectors' parameters; otherwise, they can reduce costs by updating only the weight assignor.

### 6.5 Other Experiments

While this study focuses on a text-based task, some design components also apply to other data types, such as images (Appendix V). We tested an alternative defense strategy, i.e., adversarial sample detection, but found it less effective (Appendix W). In addition, we tested the computational efficiency of our detector and found it to be practical for real-world use (Appendix X). Finally, we conducted a case study to analyze how our detector performs on specific samples and across different attack types or topics (Appendix Y).

## 7. Discussion and Conclusion

### 7.1 Research Implication

Following the computational design science paradigm (Abbasi et al. 2024, Padmanabhan et al. 2022), we develop a harmful online content detector (ARHOCD) with a LLM-SGA framework and three design components, enhancing adversarial robustness in generalizability and accuracy. We contribute prescriptive knowledge to IS research in social media management, AI security, and computational design science.

#### 7.1.1 Social Media Management in IS

Effectively managing social media to harness its benefits while mitigating adverse effects is a central topic in IS. Among various adverse effects, harmful content has received growing attention. Most studies focus on developing IT artifacts to improve the detection accuracy under safe, non-adversarial conditions (Etudo and Yoon 2024, Wei et al. 2022). However, the accuracy often drops sharply under adversarial conditions, which are common given the adversarial nature of this task. This study proposes an adversarially robust detector, emphasizing robustness as a core design dimension in these studies. Moreover, robust detection is essential for studies analyzing the mechanisms or user behaviors related to harmful content. For instance, Wu et al. (2021) first use an ML-based detector to identify online harmful content (e.g., spam), then identify the users who posted it, and finally analyze the effects of different intervention strategies on mitigating such behavior. Hence, the validity of such studies largely depends on detection accuracy because a compromised detection may misidentify users and lead to a biased conclusion. By highlighting adversarial robustness and providing a principled approach to achieve it, this study also supports this line of research.

#### 7.1.2 AI Security in IS

With the rapid advancement of AI, AI security has become a critical topic in IS. Adversarial attacks pose major threats and have drawn growing attention. This study advances IS research on AI security by proposing a framework and design components that improve the generalizability and accuracy of defenses against such attacks. Our study also offers insights into other AI security issues, such as privacy. A classic privacy threat is the membership inference attack (MIA), which infers whether a specific sample was in the training set by exploiting a model's behavioral differences between training and unseen data (Shokri et al. 2017). Our study mitigates this threat in two ways. First, our weight assignment method can be adapted to assign higher weights to base models that better protect data privacy, making MIA more difficult. This requires updating the prior from reflecting adversarial robustness to reflecting privacy-preserving ability. Second, our training set combines raw and generated samples, which protects privacy in two aspects: (1) the enlarged dataset reduces overfitting, making it harder for a model to memorize samples, and (2) even if an attacker infers that a sample was used in the training, they still cannot tell whether it is a raw or generated sample, preventing confirmation of any specific sample's presence in the original training set.

**7.1.3 Design Science Research**

This study contributes to the computational design science research paradigm, exemplifying it by tackling a practical problem (detecting harmful online content under adversarial conditions) and a situated implementation (ARHOCD) to illustrate how this paradigm solves real-world challenges. From this work, we derive two design principles. First, adversarial robustness is a critical consideration when designing new IT artifacts, especially for tasks that are inherently adversarial, such as harmful content detection (as in this study), phishing detection, and hacker asset identification. This does not mean every study must invent new robustness techniques; existing ones can often be adopted directly. In such cases, compatibility matters. Our ARHOCD, including the LLM-SGA framework and the three design components, is built on off-the-shelf models without requiring structure changes, making it a principled approach for future studies seeking to enhance adversarial robustness. We also release code at https://github.com/YiLiuHFUT/ARHOCD to further support future studies. Second, ensemble learning is critical in complex cases where a single model may fall short. Our ARHOCD assigns weights to predictions from different base detectors and across samples, leveraging their complementary strengths to achieve stronger performance. Meanwhile, as shown in our case study (Appendix Y), while the overall performance differences among base detectors are small, their performance can vary greatly on specific samples. This indicates that no single model is all-purpose to handle all cases (e.g., predicting diverse samples in our study), making ensemble learning necessary.

## 7.2 Practical Implication

This study offers important practical implications for users, social media platforms, and regulators. Users' exposure to harmful content can lead to psychological distress or misconceptions (Pennycook et al. 2020). Our study enhances the robustness of harmful content detection, helping users avoid such content and thereby improving their online experience. For social media platforms, AI has been widely used to detect harmful content, but the security issues related to AI have emerged and attracted increasing attention. Our study enhances robustness against adversarial attacks and also mitigates other AI security issues such as MIA, as discussed above. Therefore, our work helps online platforms develop more trustworthy AI-based information systems for automatically detecting harmful content, creating a beneficial and wholesome public environment for society. For regulators, reliably harnessing AI to generate social good has become an urgent priority. This requires addressing AI security issues such as adversarial attacks, which have become central topics in AI governance (Zhang and Li 2020). This study provides a technical approach to counter adversarial attacks and also sheds light on other threats such as privacy inference attacks (e.g., MIA), offering valuable insights for developing more secure and trustworthy AI-based information systems.

## 7.3 Limitations and Future Directions

This study has several limitations that suggest directions for future research. First, in harmful content detection, we enhance adversarial robustness through technical solutions. While useful, technical measures alone may not be enough. Future work can explore complementary operational strategies to achieve more dependable management. Second, our focus is on adversarial attacks and does not consider other principles such as fairness and transparency. Integrating these principles is challenging due to potential trade-offs. For example, prior work has shown a tension between adversarial robustness and transparency (Chai et al. 2023). Future work can explore how to balance these principles to build more trustworthy AI-based information systems. Third, our case study shows that some adversarial attacks are harder to defend against than others. While our method demonstrates strong worst-case performance and generalizability across diverse attacks, certain attack types are more challenging than others. Future work can focus on these more complex attacks.

# Endnotes

[1]https://www.theguardian.com/australia-news/2024/oct/11/australias-spy-chief-warns-ai-will-accelerate-online-radicalisation

# References

Abbasi A, Parsons J, Pant G, Sheng ORL, Sarker S (2024) Pathways for design research on artificial intelligence. *Inf. Syst. Res.* 35(2):441–459.

Aggarwal S, Vishwakarma DK (2024) Exposing the Achilles' heel of textual hate speech classifiers using indistinguishable adversarial examples. *Expert Syst. Appl.* 254:124278.

Anguita D, Ghio A, Oneto L, Member S, Ridella S (2011) Selecting the Hypothesis Space for Improving the Generalization Ability of Support Vector Machines. *2011 Int. Jt. Conf. Neural Networks*. (IEEE), 1169–1176.

Arjovsky M, Gulrajani I, Lopez-paz D (2019) Invariant risk minimization. *arXiv Prepr. arXiv1907.02893*.

Azumah SW, Elsayed N, Elsayed Z, Ozer M, Guardia A La (2024) Deep learning approaches for detecting adversarial cyberbullying and hate speech in social networks. *AIBThings 2024*. 1–10.

Chai Y, Liang R, Samtani S, Zhu H, Wang M, Liu Y, Jiang Y (2023) Additive feature attribution explainable methods to craft adversarial attacks for text classification and text regression. *IEEE Trans. Knowl. Data Eng.* 35(12):12400–12414.

Chang G, Gao H, Yao Z, Xiong H (2023) TextGuise: Adaptive adversarial example attacks on text classification model. *Neurocomputing* 529:190–203.

Chen H, Ji Y (2022) Adversarial training for improving model robustness? Look at both prediction and interpretation. *AAAI 2022*. 10463–10472.

Chen KC, Chen CY, Li C Te (2023) Anti-disinformation: an adversarial attack and defense network towards improved robustness for disinformation detection on social media. *BigData 2023*. 5476–5484.

Dong X, Luu AT, Ji R, Liu H (2021) Towards robustness against natural language word substitutions. *ICLR 2021*.

Etudo U, Yoon VY (2024) Ontology-based information extraction for labeling radical online content using distant supervision. *Inf. Syst. Res.* 35(1):203–225.

Fazlyab M, Entesari T, Roy A, Chellappa R (2023) Certified robustness via dynamic margin maximization and improved lipschitz regularization. *NeurIPS 2023*. 34451–34464.

Gao J, Lanchantin J, Soffa M Lou, Qi Y (2018) Black-box generation of adversarial text sequences to evade deep learning classifiers. *IEEE Secur. Priv. Work.* 50–56.

De Gibert O, Perez N, García-Pablos A, Cuadros M (2018) Hate speech dataset from a white supremacy forum. *Proc. 2nd Work. Abus. Lang. Online*. 11–20.

Glenn WB (1950) Verification of forecasts expressed in terms of probability. *Mon. Weather Rev.* 78(1):1–3.

Godinić D, Obrenovic B (2020) Effects of economic uncertainty on mental health in the COVID-19 pandemic context: social identity disturbance, job uncertainty and psychological well-being model. *Int. J. Innov. Econ. Dev.* 6(1):61–74.

Goodfellow IJ, Shlens J, Szegedy C (2015) Explaining and harnessing adversarial examples. *ICLR 2015*. 1–11.

Herel D, Cisneros H, Mikolov T (2023) Preserving semantics in textual adversarial attacks. *ECAI 2023*. 1036–1043.

Hu Z, Yin JL, Chen B, Lin L, Chen BH, Liu X (2024) Meat: median-ensemble adversarial training for improving robustness and generalization. *ICASSP 2024*. 5600–5604.

Huang P, Yang Y, Jia F, Liu M, Ma F, Zhang J (2022) Word level robustness enhancement: Fight perturbation with perturbation. *AAAI 2022*. 10785–10793.

Jin D, Jin Z, Zhou JT, Szolovits P (2020) Is BERT really robust? A strong baseline for natural language attack on text classification and entailment. *AAAI 2020*. 8018–8025.

Kingma DP, Welling M (2014) Auto-encoding variational bayes. *ICLR 2014*. 1–14.

Kumbam PR, Syed SU, Thamminedi P, Harish S, Perera I, Dorr BJ (2025) Exploiting explainability to design adversarial attacks and evaluate attack resilience in hate-Speech detection models. *ICWSM 2025*. 1038–1050.

Kurakin A, Goodfellow I, Bengio S (2017) Adversarial machine learning at scale. *ICLR 2017*.

Lecuyer M, Atlidakis V, Geambasu R, Hsu D, Jana S (2019) Certified robustness to adversarial examples with differential privacy. *IEEE Symp. Secur. Priv.* 656–672.

Leshno M, Lin VY, Pinkus A, Schocken S (1993) Multilayer feedforward networks with a nonpolynomial activation function can approximate any function. *Neural networks* 6(6):861–867.

Li L, Song D, Qiu X (2023) Text adversarial purification as defense against adversarial attacks. *ACL 2023*. 338–350.

Li W, Chai Y (2022) Assessing and enhancing adversarial robustness of predictive analytics: An empirically tested design framework. *J. Manag. Inf. Syst.* 39(2):542–572.

Liu Y, Yang C, Li D, Ding J, Jiang T (2024) Defense against adversarial attacks on no-reference image quality models with gradient norm regularization. *CVPR 2024*. 25554–25563.

Mollas I, Chrysopoulou Z, Karlos S, Tsoumakas G (2022) ETHOS: a multi-label hate speech detection dataset. *Complex Intell. Syst.* 8(6):4663–4678.

Moraffah R, Khandelwal S, Bhattacharjee A, Liu H (2024) Adversarial text purification: A large language model approach for defense. *PAKDD 2024*. 65–77.

Morris JX, Lifland E, Lanchantin J, Ji Y, Qi Y (2020) Reevaluating adversarial examples in natural language. *EMNLP 2020*. 3829–3839.

Mozes M, Stenetorp P, Kleinberg B, Griffin LD (2021) Frequency-guided word substitutions for detecting textual adversarial examples. *Proc. 16th Conf. Eur. chapter Assoc. Comput. Linguist.* 171–186.

Nguyen DM, Tuan LA (2022) Textual manifold-based defense against natural language adversarial examples. *EMNLP 2022*:6612–6625.

Oak R (2019) Poster: Adversarial examples for hate speech classifiers. *Proc. 2019 ACM SIGSAC Conf. Comput. Commun. Secur.* 2621–2623.

Ocampo NB, Cabrio E, Villata S (2023) Playing the part of the sharp bully: Generating adversarial examples for implicit hate speech detection. *ACL 2023*. 2758–2772.

Padmanabhan B, Fang X, Sahoo N, Burton-jones A (2022) Editor's comments: Machine learning in information systems research. *MIS Q.* 46(1):iii–xix.

Papernot N, McDaniel P, Swami A, Harang R (2016) Crafting adversarial input sequences for recurrent neural networks. *MILCOM 2016*. (IEEE), 49–54.

Pennycook G, Bear A, Collins ET, Rand DG (2020) The implied truth effect: Attaching warnings to a subset of fake news headlines increases perceived accuracy of headlines without warnings. *Manage. Sci.* 66(11):4944–4957.

Pruthi D, Dhingra B, Lipton ZC (2019) Combating adversarial misspellings with robust word recognition. *ACL 2019*. 5582–5591.

Przybyła P, McGill E, Saggion H (2025) Attacking misinformation detection using adversarial examples generated by language models. *EMNLP 2025*. 27614–27630.

Qin R, Wang L, Du X, Chen X, Yan B (2023) Dynamic ensemble selection based on deep neural network uncertainty estimation for adversarial robustness. *arXiv Prepr. arXiv2308.00346*.

Qiu S, Liu Q, Zhou S, Huang W (2022) Adversarial attack and defense technologies in natural language processing: A survey. *Neurocomputing* 492:278–307.

Rath A, Mishra D, Panda G, Satapathy SC, Xia K (2022) Improved heart disease detection from ECG signal using deep learning based ensemble model. *Sustain. Comput. Informatics Syst.* 35:100732.

Ribeiro MT, Singh S, Guestrin C (2016) "Why should i trust you?" Explaining the predictions of any classifier. *SIGKDD 2016*. 1135–1144.

Sen S, Ravindran B, Raghunathan A (2020) EMPIR: Ensembles of mixed precision deep networks for increased robustness against adversarial attacks. *ICLR 2020*.

Shokri R, Stronati M, Song C, Shmatikov V (2017) Membership inference attacks against machine learning models. *IEEE Symp. Secur. Priv.* 3–18.

Ting KM (2002) An instance-weighting method to induce cost-sensitive trees. *IEEE Trans. Knowl. Data Eng.* 14(3):659–665.

Waghela H, Sen J, Rakshit S, Dasgupta S (2024) Adversarial robustness through dynamic ensemble learning. *SILCON 2024*. 1–6.

Wang J, Bao R, Zhang Z, Zhao H (2022) Rethinking textual adversarial defense for pre-trained language models. *IEEE/ACM Trans. Audio Speech Lang. Process.* 30:2526–2540.

Waseda F, Chang C chun, Echizen I (2025) Rethinking invariance regularization in adversarial training to improve robustness-accuracy trade-off. *ICLR 2025*.

Wei X, Zhang Z, Zhang M, Chen W, Zeng DD (2022) Combining crowd and machine intelligence to detect false news on social media. *MIS Q.* 46(2):977–1008.

Wu J, Zheng Z, Zhao JL (2021) FairPlay: Detecting and deterring online customer misbehavior. *Inf. Syst. Res.* 32(4):1323–1346.

Yang Y, Liu X, He K (2025) Synonym-unaware fast adversarial training against textual adversarial attacks. *NAACL 2025*. 727–739.

Yang Y, Wang X, He K (2022) Robust textual embedding against word-level adversarial attacks. *UAI 2022*. 2214–2224.

Ye M, Gong C, Liu Q (2020) SAFER: A structure-free approach for certified robustness to adversarial word substitutions. *ACL 2020*. 3465–3475.

Yoo JY, Qi Y (2021) Towards improving adversarial training of NLP models. *EMNLP 2021*. 945–956.

Zeng J, Xu J, Zheng X, Huang X (2023) Certified robustness to text adversarial attacks by randomized [MASK]. *Comput. Linguist.* 49(2):395–427.

Zhang C, Zhou X, Wan Y, Zheng X, Chang KW, Hsieh CJ (2022) Improving the adversarial robustness of NLP models by information bottleneck. *ACL 2022*. 3588–3598.

Zhang J, Li C (2020) Adversarial examples: opportunities and challenges. *IEEE Trans. Neural Networks Learn. Syst.* 31(7):2578–2593.

Zhang S, Gao H, Rao Q (2021) Defense against adversarial attacks by reconstructing images. *IEEE Trans. Image Process.* 30:6117–6129.

Zhang X, Hong H, Hong Y, Huang P, Wang B, Ba Z, Ren K (2024) Text-crs: A generalized certified robustness framework against textual adversarial attacks. *Proc. - Secur. Priv.*:2920–2938.

Zhang Y, Yang Q (2018) An overview of multi-task learning. *Natl. Sci. Rev.* 5(1):30–43.

Zhao J, Wei P, Mao W (2021) Robust neural text classification and entailment via mixup regularized adversarial training. *SIGIR 2021*. 1778–1782.

Zhou Y, He B, Sun L (2024) Humanizing machine-generated content: evading AI-text detection through adversarial attack. *Lr. 2024*. 8427–8437.

Zhou Y, Zheng X, Hsieh CJ, Chang KW, Huang X (2021) Defense against synonym substitution-based adversarial attacks via dirichlet neighborhood ensemble. *ACL-IJCNLP 2021*. 5482–5492.